\documentclass{article} 
\usepackage[preprint]{colm2025_conference}

\usepackage{microtype}
\usepackage{hyperref}
\usepackage{url}
\usepackage{booktabs}
\usepackage{todonotes}
\usepackage{xcolor}
\usepackage[most]{tcolorbox}
\usepackage{lineno}
\usepackage{xspace}
\usepackage{amsmath}
\usepackage{graphicx}
\usepackage{subcaption}
\usepackage{multirow}
\usepackage{adjustbox}
\usepackage{pgfplots}
\usepackage{multicol}
\pgfplotsset{compat=1.18} 
\usepackage{enumitem}
\usepackage{soul}
\usepackage{wrapfig}
\usepackage{array}
\usepackage{fancyvrb}
\usepackage[utf8]{inputenc}
\usepackage{amssymb}  

\definecolor{darkblue}{rgb}{0, 0, 0.5}
\hypersetup{colorlinks=true, citecolor=darkblue, linkcolor=darkblue, urlcolor=darkblue}

\title{
    Reasoning Models Can Be Effective Without Thinking 
}

\author{Wenjie Ma$^1$ \quad 
Jingxuan He$^1$ \quad 
Charlie Snell$^1$ \quad 
Tyler Griggs$^1$ \\
\textbf{Sewon Min$^{1,2}$ \quad 
Matei Zaharia$^1$}
\\
\textsuperscript{1}University of California, Berkeley \quad
\textsuperscript{2}Allen Institute for AI \\
\texttt{\{windsey,jingxuan.he,csnell22,tgriggs,sewonm,matei\}@berkeley.edu}
}

\newcommand{\myskip}[1]{}

\newcommand{\myhl}[2]{{%
    \sethlcolor{#1}%
    \hl{#2}%
    }%
}
\definecolor{overview-thinking}{RGB}{194, 213, 247}
\definecolor{overview-nothinking}{RGB}{252, 225, 198}
\definecolor{exp-thinking}{RGB}{29, 108, 171}
\definecolor{exp-nothinking}{RGB}{255, 116, 16}

\newcommand{\thinking}{\textit{Thinking}\xspace}
\newcommand{\nothinking}{\textit{NoThinking}\xspace}
\newcolumntype{P}[1]{>{\raggedleft\arraybackslash}p{#1}}

\newif\ifshowcomments
\showcommentsfalse

\ifshowcomments
    \newcommand{\wj}[1]{\textcolor{blue}{[Wenjie: #1]}}
    \newcommand{\sm}[1]{{\textcolor{violet!80!black}{[Sewon: #1]}}}
    \newcommand{\jh}[1]{\textcolor{brown}{[Jingxuan: #1]}}
\else
    \newcommand{\wj}[1]{}
    \newcommand{\sm}[1]{}
    \newcommand{\jh}[1]{}
\fi

\begin{document}

\ifcolmsubmission
\linenumbers
\fi

\maketitle

\begin{abstract}
Recent LLMs have significantly improved reasoning capabilities, primarily by including an explicit, lengthy \thinking process as part of generation.
In this paper, we question whether this explicit thinking is necessary.
Using the state-of-the-art DeepSeek-R1-Distill-Qwen, we find that bypassing the thinking process via simple prompting, denoted as \nothinking, can be surprisingly effective. When controlling for the number of tokens, \nothinking outperforms \thinking across a diverse set of seven challenging reasoning datasets---including mathematical problem solving, formal theorem proving, and coding---especially in low-budget settings, e.g.,  51.3 vs. 28.9 on ACM 23 with 700 tokens. 
Notably, the performance of \nothinking becomes more competitive with pass@$k$ as $k$ increases. 
Building on this observation, we demonstrate that a parallel scaling approach that uses \nothinking to generate $N$ outputs independently and aggregates them is highly effective. For aggregation, we use task-specific verifiers when available, or we apply simple best-of-$N$ strategies such as confidence-based selection.
Our method outperforms a range of baselines with similar latency using \thinking, and is comparable to \thinking with significantly longer latency (up to 9x).
Together, our research encourages a reconsideration of the necessity of lengthy thinking processes, while also establishing a competitive reference for achieving strong reasoning performance in low-budget settings or at low latency using parallel scaling.

\end{abstract}

\section{Introduction}
\label{sec:intro}
\begin{figure}[t]
    \begin{minipage}{.66\textwidth}
        \centering
        \includegraphics[width=\textwidth]{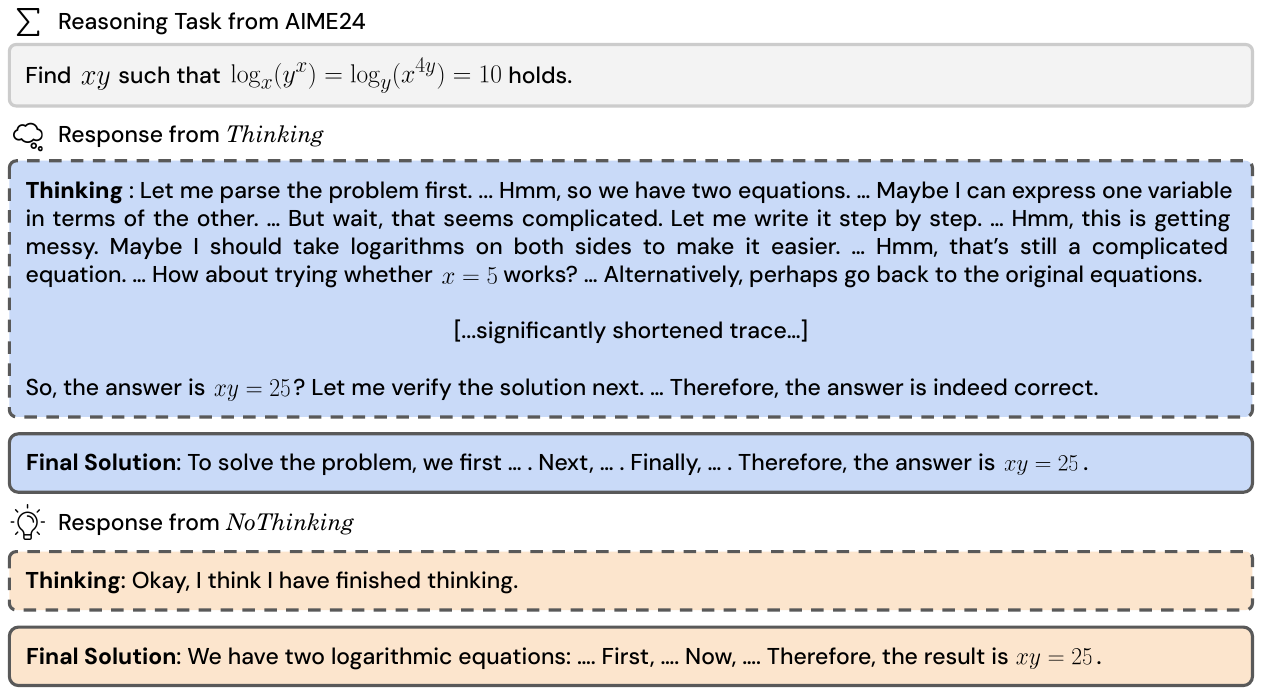}
        \vspace{-5mm}
        \caption{Before producing the final step-wise solution, \myhl{overview-thinking}{\textit{Thinking}} generates a long chain of thought in the dedicated \emph{thinking box}, containing a search process with reflection, backtracking, and self-verification. In contrast, \myhl{overview-nothinking}{\textit{NoThinking}} directly generates the solution after a prefilled dummy thinking box.}
        \label{fig:overview}
    \end{minipage}
    \hfill
    \begin{minipage}{.3\textwidth}
        \begin{minipage}{\textwidth}
            \centering
            \includegraphics[height=24mm]{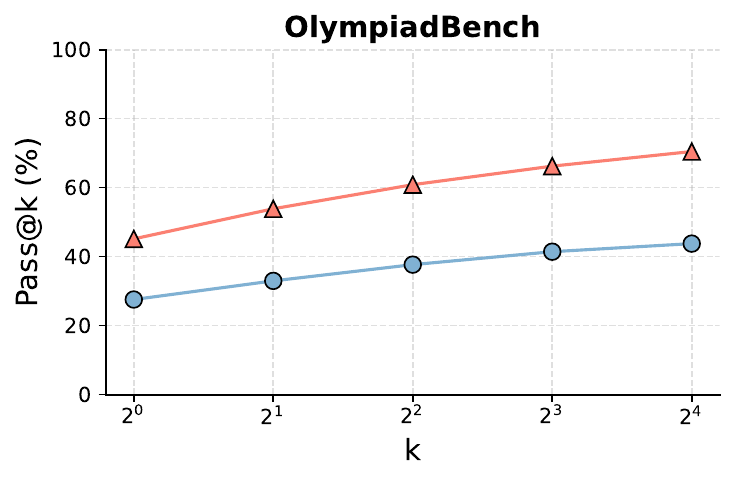}
            \vspace{-3.5mm}
            \caption{\textcolor{exp-nothinking}{\nothinking} outperforms \textcolor{exp-thinking}{\thinking} at a similar budget of 800 tokens.}
            \label{fig:overview-exp1}
            \vspace{2mm}
        \end{minipage}
        \begin{minipage}{\textwidth}
            \centering
            \includegraphics[height=24mm]{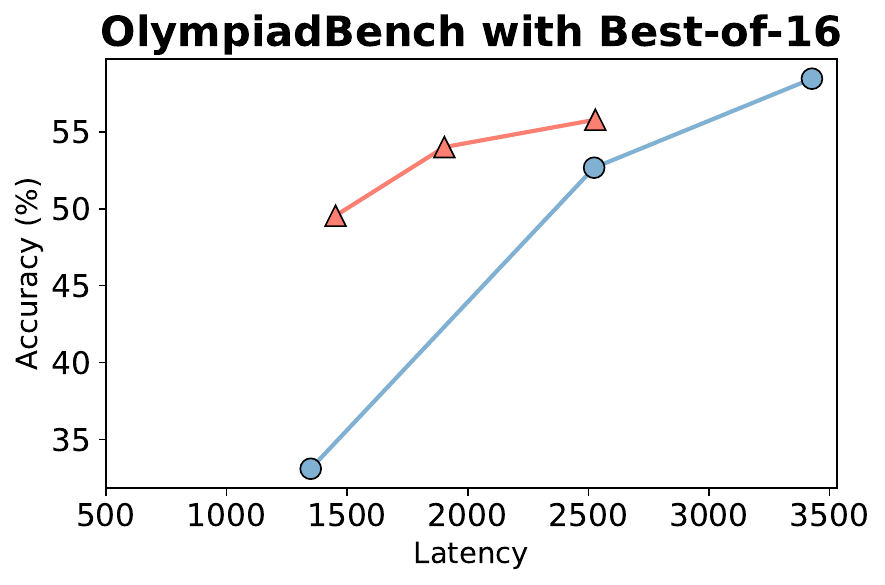}
            \vspace{-3.2mm}
            \caption{When controlling for latency, \textcolor{exp-nothinking}{\nothinking} significantly outperforms \textcolor{exp-thinking}{\thinking}. Results are shown with best-of-$N$ sample selection applied.}
            \vspace{1mm}
            \label{fig:overview-exp2}
        \end{minipage}
    \end{minipage}
\end{figure}

Recent large reasoning models, such as DeepSeek-R1~\citep{deepseek-r1}, OpenAI o1~\citep{openai-o1}, and others~\citep{qwen2025qwq32b,simplescaling,sky-t1,ye2025limoreasoning}, have significantly improved performance by increasing inference-time compute~\citep{snell2024scalingllmtesttimecompute,brown2024largelanguagemonkeysscaling}. As shown in~\autoref{fig:overview}, these models approach complex tasks by first generating long chains of thought as a search process with reflection, backtracking, and self-validation (``\textbf{Thinking}'', upper half of the blue box), followed by final solution steps and a final answer (``\textbf{Final Solution}'', lower half). These reasoning abilities are typically acquired through reinforcement learning with verified rewards or fine-tuning on distilled reasoning traces. These models are generally believed to benefit from this explicit reasoning process, which has led to the popular paradigm of inference-time compute scaling—enhancing reasoning at the cost of significantly increased token usage and latency.
Recent work has explored efficient reasoning by reducing the token usage during the thinking process, e.g., via reinforcement learning with length-based rewards~\citep{aggarwal2025l1controllinglongreasoning} or supervised fine-tuning on variable-length CoT traces~\citep{efficientreasoningsurvey}.
However, most approaches still rely on the explicit \thinking process.

In this paper, we question whether the explicit \thinking process is necessary for advanced reasoning. With an established reasoning model, DeepSeek-R1-Distill-Qwen, we disabled the explicit thinking process through a simple prompting approach (orange box in Figure~\ref{fig:overview}). Our approach, called \nothinking, simply prefills the assistant response with a fabricated \thinking block and has the model continue from there.
We evaluate this approach extensively on diverse benchmarks, including math problem solving (AIME 2024, AIME 2025, AMC 2023)~\citep{olympiadbench}, coding~\citep{jain2025livecodebench}, and formal theorem proving~\citep{zheng2022miniff,proofnet}.

Our experimental results show that \nothinking is surprisingly effective, as evidenced by pass@$k$ metrics ($1 \leq k \leq 64$).
Out of the box, \nothinking, using 2.0–5.1× fewer tokens, either matches \thinking across all values of $k$ or initially lags at $k=1$ but quickly catches up and sometimes surpasses \thinking as $k$ increases. This pattern is not observed in the base models used to train the reasoning model.
Moreover, when controlling the token usage of both approaches through the budget forcing technique~\citep{simplescaling},
\nothinking often achieves higher pass@1 accuracy than \thinking in the low-budget regime, with the performance gap expanding as $k$ increases.
This is exemplified in Figure~\ref{fig:overview-exp1}.
To systematically assess efficiency, we examine the Pareto frontier of pass@$k$ versus average token usage, and find that \nothinking consistently outperforms \thinking.
This suggests that even without the explicit thinking process, \nothinking\ achieves significantly better accuracy-cost tradeoffs across a wide range of budgets compared to the default prompting with \thinking.

The advantage of \nothinking in pass@$k$ is so significant that it can enable new inference strategies that achieve better accuracies and latencies than sequential inference with \thinking.
To illustrate, we explore implementing parallel scaling with \nothinking, where multiple responses are sampled in parallel and aggregated via best-of-$N$ methods~\citep{kang2025scalablebestofnselectionlarge,wang2023selfconsistencyimproveschainthought}.
We consider both tasks with perfect verifiers (e.g., formal theorem proving), where correctness can be automatically verified, and tasks without verifiers (e.g., general problem solving), where we rely on simple confidence-based selection strategies. For tasks with verifiers, \nothinking with parallel scaling outperforms \thinking, both with and without parallel scaling, achieving 7× lower latency and even 4× less total token usage. For some tasks without verifiers, \nothinking also demonstrates strong performance: for example, it surpasses \thinking with 9× lower latency and improved accuracy on OlympiadBench (Math). As shown in~\autoref{fig:overview-exp2}, we observe a consistent pattern across benchmarks: \nothinking with parallel scaling offers a superior trade-off between accuracy and latency.
This illustrates that, for many tasks, reasoning model providers may be able to deliver better latency and similar or improved quality through similar approach.

To summarize, our work is the first to show the surprising finding that the thinking process of current reasoning models is not necessarily required for high performance, even for models that have been trained to follow this structured reasoning format. Moreover, we show that this simple \nothinking prompting can be combined with parallel test-time compute to provide better  latency-vs-accuracy than sequential inference. Overall, our work encourages a reconsideration of the necessity of lengthy thinking processes, and provides a competitive baseline for strong reasoning performance in low-budget or low-latency settings.



\section{Related Work and Background} 
\label{sec:related_work}

\paragraph{Test-Time Scaling for Language Models} Scaling inference-time compute has emerged as an effective strategy to enhance the reasoning capabilities of large language models (LLMs). Broadly, existing methods fall into two categories. The first is \textbf{sequential} approaches exemplified by models such as OpenAI’s o1, DeepSeek R1, and Qwen QwQ~\citep{deepseek-r1, openai-o1, qwen2025qwq32b}. These models produce long, structured chain-of-thought (CoT) responses involving backtracking and verification in a single forward pass. To gain such capabilities, costly training procedures—such as reinforcement learning (RL) or iterative self-improvement—are required to bootstrap and refine reasoning ability~\citep{zelikman2022step, lambert2025tulu3, yuan2024freeprocessrewardsprocess}. For instance, DeepSeek R1 is trained entirely via RL to generate extended CoTs, while RLVR~\citep{lambert2025tulu3} and PRIME~\citep{yuan2024freeprocessrewardsprocess} leverage verifiable or latent supervision to improve reasoning quality. The second category is \textbf{parallel} approaches that sample multiple outputs and aggregate them. These include Best-of-$N$ sampling and search-guided decoding techniques such as Monte Carlo Tree Search (MCTS)~\citep{snell2024scalingllmtesttimecompute, brown2024largelanguagemonkeysscaling}, where multiple candidate outputs are generated and selectively aggregated using search algorithms, reward models, or external verifiers.

In contrast to prior work that assumes explicit thinking processes are essential for strong performance, our work demonstrates that disabling explicit thinking is promising to improve the accuracy-budget tradeoff. Our prompting-based method, simple yet overlooked, achieves competitive performance without requiring additional training, reward signals, or process supervision. While we utilize Best-of-$N$ methods to illustrate the effectiveness of our approach, our goal is not to innovate on sampling itself, but rather to present a strong, cost-effective baseline for low-budget settings.

\paragraph{Efficient Reasoning} Given the growing compute cost of sequential scaling in reasoning models, recent work has explored various strategies to make reasoning in LLMs more efficient. These approaches aim to promote smart and concise reasoning by optimizing the length of generated reasoning sequences and eliminating unnecessary steps, as summarized in a recent survey~\citep{efficientreasoningsurvey}. Some methods continue to train reasoning models for full-length CoT optimization, such as applying reinforcement learning with explicit length-based rewards~\citep{aggarwal2025l1controllinglongreasoning,luo2025o1prunerlengthharmonizingfinetuningo1like,shen2025dastdifficultyadaptiveslowthinkinglarge,arora2025traininglanguagemodelsreason,qu2025optimizingtesttimecomputemeta}, or fine-tuning on variable-length or concise reasoning traces obtained from best-of-N sampling~\citep{xia2025tokenskipcontrollablechainofthoughtcompression,kang2024c3otgeneratingshorterchainofthought,ma2025cotvalvelengthcompressiblechainofthoughttuning,munkhbat2025selftrainingelicitsconcisereasoning,yu2024distilling21,liu2024languagemodelslearnskip,cui2025stepwiseperplexityguidedrefinementefficient}. Other approaches modify the output paradigm itself to promote concise reasoning, for example by training LLMs to leverage latent representations~\citep{coconut, cheng2024compressedchainthoughtefficient,shen2025efficientreasoninghiddenthinking,xu2025softcotsoftchainofthoughtefficient,su2025tokenassortedmixinglatent,saunshi2025reasoninglatentthoughtspower,shen2025codicompressingchainofthoughtcontinuous}, or adopting training-free techniques that select appropriate criteria to guide the reasoning strategy~\citep{fu2024efficientlyservingllmreasoning,wu2025lessunderstandingchainofthoughtlength,aytes2025sketchofthoughtefficientllmreasoning,liao2025rewardguidedspeculativedecodingefficient,sun2024fastbestofndecodingspeculative,li2025fastmctssimplesamplingstrategy,ding2025dynamicparalleltreesearch,wang2025samplingefficienttesttimescalingselfestimating,zhang2025lightthinkerthinkingstepbystepcompression,yan2025inftythinkbreakinglengthlimits}. Additionally, some methods explicitly instruct LLMs to generate fewer reasoning steps~\citep{han2025tokenbudgetawarellmreasoning,xu2025chaindraftthinkingfaster,lee2025llmscompresschainofthoughttoken,Renze_2024,thinkingintervention} (e.g., setting a token
budget in prompts), or dynamically route inputs based on task difficulty to control reasoning complexity~\citep{ong2025routellmlearningroutellms,chuang2025learningroutellmsconfidence,chuang2025confidentseekstrongerexploring}. While some of these methods are effective, our approach is entirely orthogonal: we show that disabling the thinking process can be surprisingly effective.
\section{\nothinking\ Provides Better Accuracy-budget Tradeoffs than \thinking}\label{sec:main-exp}

In this section, we show that \nothinking\ can be surprisingly effective, outperforming \thinking with budget forcing and offering better accuracy-budget tradeoffs. We first define \thinking and \nothinking (Section~\ref{subsec:method}), describe our experimental setup (Section~\ref{subsec:eval-setup}), and present experimental results (Section~\ref{subsec:main-results}).

\subsection{Method}\label{subsec:method}

Most modern reasoning models, such as R1 and R1-Distill-Qwen, follow a similar structure in their generation process: the reasoning process within the thinking box, marked by \texttt{<|beginning\_of\_thinking|>} and \texttt{<|end\_of\_thinking|>}, followed by the final answer.
Based on this structure, we define the two methods (\thinking and \nothinking) as follows.

\vspace{.2em} \noindent
\textbf{\thinking} refers to the default method of querying the reasoning models to produce the following outputs: the reasoning process within the thinking box, the final solution, and the final answer (\autoref{fig:overview} (blue)). 

\vspace{.2em} \noindent
\textbf{\nothinking} refers to a method that bypasses the explicit reasoning process through prompting, directly generating the final solution and answer. This is achieved by forcing the thinking box to be empty during the decoding process (\autoref{fig:overview} (orange)).
%
\begin{quote}
\texttt{<|beginning\_of\_thinking|>} \newline
\texttt{Okay, I think I have finished thinking.} \newline
\texttt{<|end\_of\_thinking|>}
\end{quote}


The exact prompts we use can be found in Appendix~\ref{subsec:example-outputs}.

To control token usage in both methods, we adapt a budget forcing technique from \citet{simplescaling}. When the model reaches the token budget, it is forced to generate \texttt{Final Answer:}, ensuring an immediate final answer. If the model is still within the thinking box at the token limit, \texttt{<|end\_of\_thinking|>} is appended before the final answer tag. To implement budget forcing effectively, we slightly adjust the implementation for each task; see Appendix~\ref{app:budget-forcing-details} for details.


\subsection{Evaluation Setup}\label{subsec:eval-setup}

\paragraph{Models}  We use \textbf{DeepSeek-R1-Distill-Qwen-32B}~\citep{deepseek-r1} as our primary model. It is a distilled variant of DeepSeek-R1, created by initializing the model with Qwen-32B and training on the data generated by DeepSeek-R1.
It is one of the state-of-the-art reasoning models that use sequential test-time scaling, and is on par with a larger, DeepSeek-R1-Distill-Llama-70B according to the report.
As an additional baseline, we also include \textbf{Qwen-32B-Instruct}~\citep{qwen2025qwen25technicalreport}---a general-purpose instruction-tuned model that uses the same architecture but was not specifically designed for structured reasoning. We have also experimented with R1-series models at smaller scales (7B and 14B) on AIME tasks, and both exhibit similar behavior to the 32B model used in our main experiments. The results have been added to Appendix~\ref{sec:more_models}. We lack access for making interventions to closed-source models.
It is left to future work to explore models from other providers that follow the Thinking-Solution format introduced in Section~\ref{sec:intro}. \wj{TODO: Mention closed models?}

\paragraph{Tasks and Benchmarks}

We include a diverse set of challenging reasoning benchmarks, covering tasks such as competitive math, coding, Olympiad problems, and theorem proving.

\begin{itemize}[leftmargin=16pt, topsep=1pt]
    \item Mathematical problem solving: We include both standard and Olympiad-level math tasks. For standard problem solving, we use \textbf{AIME 2024}, \textbf{AIME 2025}, and \textbf{AMC 2023}, which are widely used benchmarks for reasoning models~\citep{deepseek-r1,openai-o1}. For more advanced reasoning, we include the math subset of \textbf{OlympiadBench}~\citep{olympiadbench}, which is much more challenging than AIME and AMC.
    
    \item Coding: We include \textbf{LiveCodeBench}~\citep{jain2025livecodebench} (release\_v2), a holistic and contamination-free benchmark that continuously collects new problems over time. 
    
    
    \item Formal theorem proving: We include \textbf{MiniF2F}~\citep{zheng2022miniff} for formal mathematical reasoning, and \textbf{ProofNet}~\citep{proofnet} for logic and theorem proving. These two datasets are among the most commonly used benchmarks for theorem proving in formal languages.
    
\end{itemize}


\paragraph{Metrics}
We report pass@$k$, which measures the probability of obtaining at least one correct output among $k$ randomly selected samples out of $n$ generated completions per problem. It is formally defined as $\text{pass@}k = \mathbb{E}_{\text{problems}} \left[ 1 - \frac{\binom{n - c}{k}}{\binom{n}{k}} \right]$, where $n$ is the number of sampled outputs per problem, and $c$ is the number of correct outputs among them~\citep{chen2021evaluatinglargelanguagemodels}.
We used $k=\{1,2,4,8,16,32\}$ for theorem proving datasets (MiniF2F and ProofNet) following the standard, $k=\{1,2,4,8,16,32,64\}$ for smaller datasets (AIME24, AIME25, AMC23) and $k=\{1,2,4,8,16\}$ for larger datasets (OlympiaddBench, LiveCodeBench). For formal theorem-proving benchmarks, $\text{pass@}32$ is the standard, while for math and coding benchmarks, $\text{pass@}1$ (i.e., accuracy) is most commonly used. 

\begin{figure}[t]
    \centering
    \includegraphics[width=\textwidth]{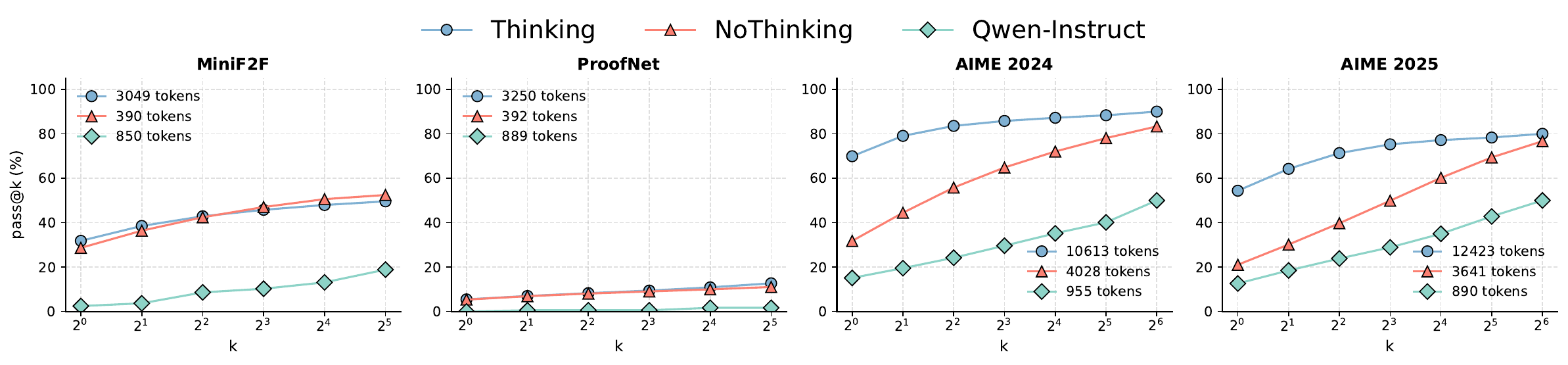}
    \includegraphics[width=0.77\textwidth]{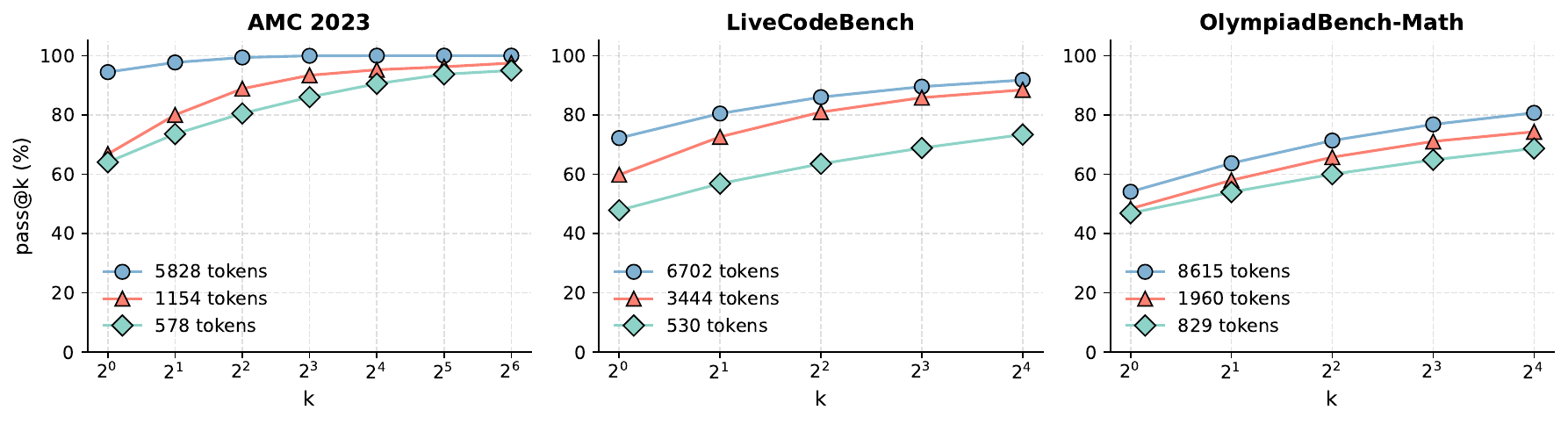}

  \caption{\textbf{\thinking vs.~\nothinking vs.~Qwen-Instruct \emph{without} token budget controlled.} 
    Blue, orange, and green indicate \thinking, \nothinking, and Qwen-Instruct, respectively. The legend in each subfigure indicates the average number of tokens used by each approach.
  }
    \label{fig:no_budget_forcing}
\end{figure}

\subsection{Results}\label{subsec:main-results}

\begin{figure}[t]
  \centering
  \includegraphics[width=\textwidth]{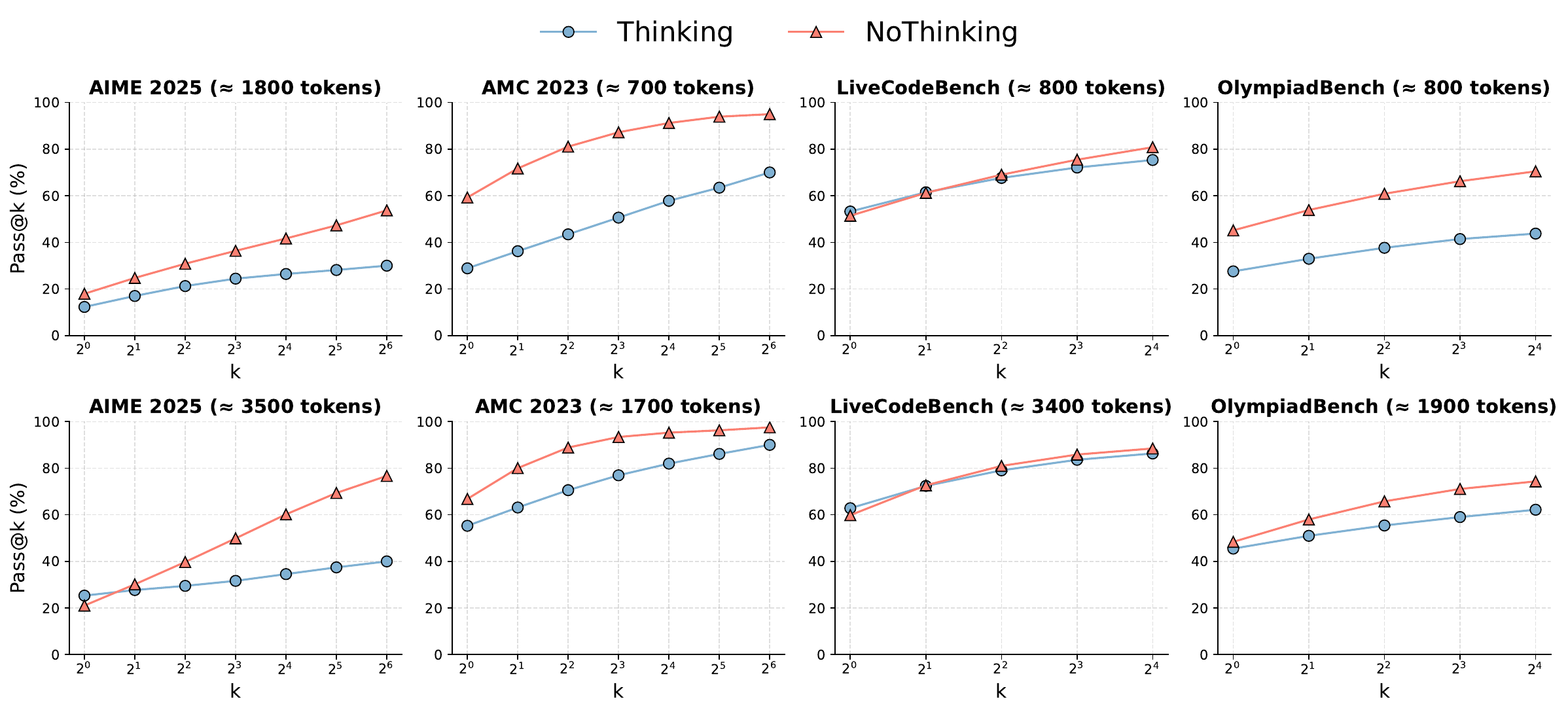}

  

  \caption{
  \textbf{\thinking vs. \nothinking~\emph{with} token budget controlled.} Orange and blue indicate \nothinking and \thinking, respectively. Under comparable token budget, \nothinking consistently outperforms \thinking as $k$ increases.}
  \label{fig:passk_similar_budget}
\end{figure}

\begin{figure}[t]
    \centering
    \includegraphics[width=\textwidth]{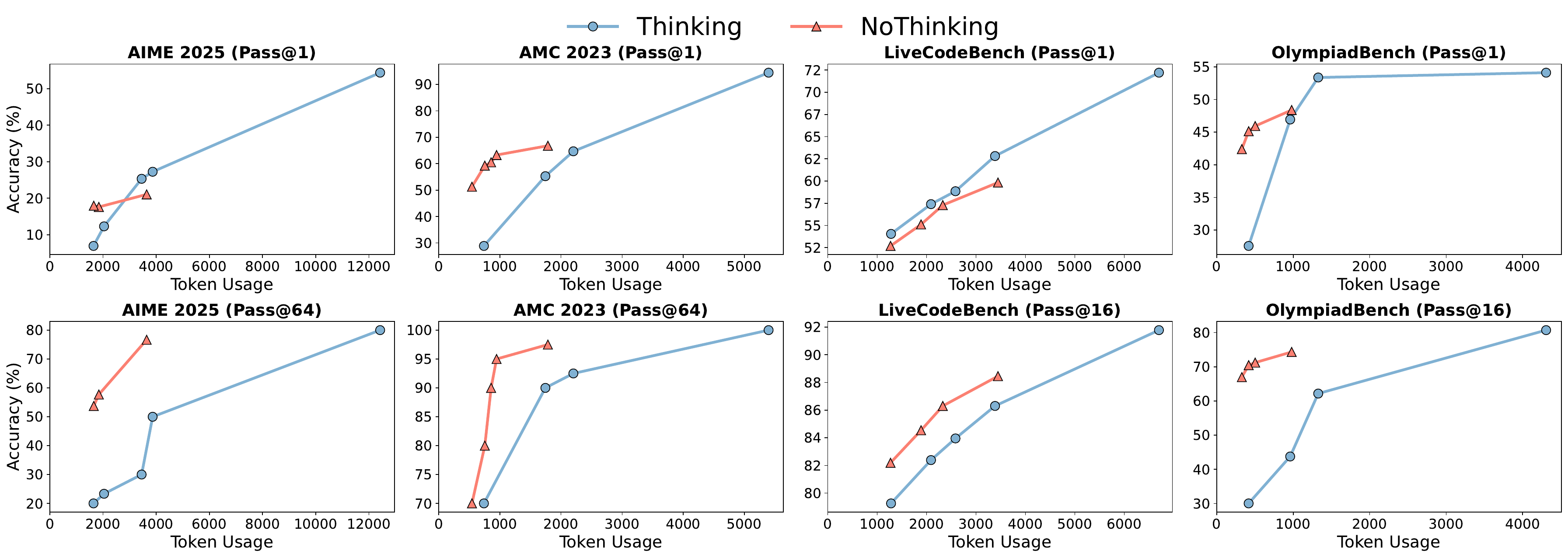}



    \caption{Pass@$k$ versus average token usage across different benchmarks. Each curve represents a different method: blue for \thinking and orange for \nothinking. We use both $k=1$ and the largest available $k$ for each benchmark.}
    \label{fig:pareto}
\end{figure}


\paragraph{\thinking vs.~\nothinking vs.~Qwen Instruct \emph{without} token budget controlled}
\autoref{fig:no_budget_forcing} compares three models without budget forcing 
across different values of pass@$k$.

First, on MiniF2F and ProofNet, \nothinking is comparable to \thinking across all values of $k$ (both significantly outperforming Qwen-Instruct). This is particularly surprising, given that \nothinking is using 3.3--3.7x fewer tokens than \thinking. On other datasets, results are more mixed. At $k=1$, \nothinking lags behind \thinking, but the gap narrows as $k$ increases. Across all datasets, \nothinking eventually matches \thinking at the largest $k$, while still using 2.0–5.1x fewer tokens.
For AIME24, AIME25, and LiveCodeBench, both \thinking and \nothinking significantly outperform Qwen-Instruct. However, on AMC23 and OlympiadBench, Qwen-Instruct also closes the gap with \thinking and \nothinking. Example inputs and outputs for different tasks and approaches are provided in Appendix~\ref{subsec:example-outputs}.

\paragraph{\thinking vs. \nothinking \emph{with} token budget controlled}
\nothinking lags behind \thinking on datasets other than MiniF2F and ProofNet, but it also uses significantly fewer tokens. Therefore, we next compare \thinking and \nothinking with similar token budget on these benchmarks, controlled by budget forcing described in Section~\ref{subsec:method}.

As shown in \autoref{fig:passk_similar_budget}, when token usage is controlled, \nothinking generally outperforms \thinking. In particular, under a low-budget setting (e.g., fewer than $\approx3,000$ tokens), \nothinking consistently achieves better results than \thinking across all values of $k$, with the performance gap widening as $k$ increases. When token usage is higher (e.g., around $3,500$ tokens), \thinking performs better than \nothinking at pass$@1$, but \nothinking quickly surpasses \thinking starting from $k=2$.

\autoref{fig:pareto} further illustrates these results by plotting token usage on the x-axis, comparing both pass$@1$ and pass$@k$ with the largest available $k$. With pass$@k$, \nothinking consistently outperforms \thinking across the entire budget range, without exception. For pass$@1$, \nothinking is better in the low-budget regime, while worse in the high-budget regime. LiveCodeBench is an exception, where \thinking outperforms \nothinking even in the low-budget regime at pass$@1$. This is likely due to the fact that on LiveCodeBench, disabling the thinking box does not significantly reduce token usage.

To mitigate the risk of data contamination, we include the newly released AIME 2025 which is unlikely to have appeared in the pretraining data of existing models. Importantly, our findings remain consistent across both new and established benchmarks, suggesting that the observed trends are not artifacts of memorization but instead reflect generalizable model behavior.

\paragraph{Summary} Despite removing the thinking box---a core component of modern reasoning models---\nothinking remains surprisingly effective. It matches \thinking while using 3.3–3.7x fewer tokens on theorem-proving benchmarks and outperforms \thinking with a similar token budget on other benchmarks, especially in low-budget settings---for instance, 51.3 vs. 28.9 on AMC23 with 700 tokens.
Furthermore, \nothinking consistently outperforms \thinking across the entire budget range for pass$@k$ with $k > 1$, with its advantage growing as $k$ increases.

\subsection{Discussions and Analyses}
\paragraph{Task-Specific Differences in \nothinking Performance} Despite the consistent trends presented in Section~\ref{subsec:main-results}, individual benchmarks reveal nuanced differences in how \nothinking performs. In~\autoref{fig:no_budget_forcing}, AMC 2023 shows a notable convergence across all three settings, with minimal performance gaps—potentially suggesting saturation on this benchmark. Moreover, MiniF2F and ProofNet exhibit surprising behavior: \nothinking achieves accuracy comparable to \thinking at pass@1, while using significantly fewer tokens. However, this should not be interpreted as task simplicity. In fact, our evaluation shows that even a strong model like OpenAI’s o1 achieves only 30\% accuracy on MiniF2F, and ProofNet consistently yields low scores across all methods. These results emphasize the importance of considering task-specific variation in \nothinking effectiveness and that the strong performance of \nothinking on certain benchmarks remains an open question for future work.

\paragraph{How Increasing $k$ Affects \nothinking Performance} We briefly explore potential reasons for the observed changes in \nothinking performance as k increases, focusing on the diversity of generated answers. We measure diversity by computing the entropy of the answer distribution for each question. Specifically, for a question with answer distribution \( \{p_1, p_2, \dots, p_n\} \), the entropy is defined as
\[
H = -\sum_{i=1}^{n} p_i \log p_i,
\]
where \( p_i \) is the empirical probability of the \(i\)-th unique answer. We then summarize diversity using the \emph{mean} and \emph{standard deviation} of entropy across all questions. A higher mean entropy indicates greater overall diversity, while a lower standard deviation suggests more consistent diversity across questions. Our analysis is based on the \thinking vs. \nothinking comparisons presented in~\autoref{fig:passk_similar_budget}, excluding LiveCodeBench, which lacks exact answers.

\begin{table}[h]
\centering
\small
\begin{tabular}{lccc}
\toprule
\textbf{Task} & \textbf{Token Budget} & \textbf{\thinking} & \textbf{\nothinking} \\
\midrule
\multirow{2}{*}{AIME 2025} 
  & 1800 & 3.66 ($\pm$1.59) & 3.76 ($\pm$1.42) \\
  & 3500 & 3.03 ($\pm$2.06) & 3.43 ($\pm$1.37) \\
\midrule
\multirow{2}{*}{AMC 2023} 
  & 700  & 1.98 ($\pm$1.66) & 1.45 ($\pm$1.34) \\
  & 1700 & 1.36 ($\pm$1.52) & 1.34 ($\pm$1.23) \\
\midrule
\multirow{2}{*}{OlympiadBench} 
  & 800  & 1.63 ($\pm$1.12) & 1.48 ($\pm$1.12) \\
  & 1900 & 1.16 ($\pm$1.13) & 1.42 ($\pm$1.12) \\
\bottomrule
\end{tabular}
\caption{Comparison of mean (± std) scores under different token budgets for \thinking and \nothinking modes.}
\label{tab:entropy}
\end{table}

We observe no consistent advantage of either mode in terms of average diversity from~\autoref{tab:entropy}. In some settings, \nothinking yields higher mean entropy (AIME 2025); in others, Thinking does (AMC 2023). However, \nothinking consistently exhibits lower variance across questions, which suggests that \nothinking  generates responses with more uniform diversity across examples. We hypothesize that this increased consistency in diversity may contribute to the improved pass@$k$ performance observed with \nothinking as $k$ increases, though diversity alone does not fully explain the performance differences.

\section{
    \nothinking Makes Parallel Test-Time Compute More Effective
}
\label{sec:parallel_sampling}



Section~\ref{sec:main-exp} demonstrates that the advantage of \nothinking\ becomes more pronounced in pass@$k$ as $k$ increases. This suggests that \nothinking\ can be effectively utilized with \emph{parallel scaling methods}---generating $N$ samples independently in parallel and selecting the best-of-$N$ prediction. In this section, we show that such methods can outperform the configurations possible through \thinking in terms of accuracy vs.~latency.

\subsection{Motivation and Methods}\label{subsec:parallel-v-sequential}

\paragraph{Parallel Scaling v. Sequential Scaling} Parallel scaling inherently provides low \textbf{latency}, as multiple model invocations can be executed concurrently--either via API calls or local model serving. This can be achieved through a multi-GPU setup or through batching on a single GPU, resulting in higher GPU utilization compared to sequential scaling. The overall latency is determined by the longest individual generation time. Motivated by our findings that \nothinking\ generates more accurate solutions in the low-budget regime and becomes increasingly effective in pass@$k$ as $k$ grows, we show that, when combined with simple best-of-$N$ methods, \nothinking\ with parallel sampling significantly improves accuracy. It outperforms alternatives with comparable latency, such as \thinking\ with budget forcing and parallel sampling, and even surpasses the pass@1 performance of full \thinking\ (\thinking without budget forcing) under sequential scaling, despite incurring much lower latency.

\paragraph{Methods}

Parallel sampling requires a method that aggregates $N$ independent solutions to produce a single prediction, so-called \emph{best-of-$N$} method.
Formally, given $N$ predictions $\mathcal{P} = {p_1, \cdots, p_N}$, the best-of-$N$ method selects $p \in \mathcal{P}$ as the final output. For tasks like MiniF2F and ProofNet, we assume access to a \textbf{perfect verifier} $f$, which reliably determines whether a given prediction $p \in \mathcal{P}$ is correct--the Lean~\citep{10.1007/978-3-030-79876-5_37} compiler. For other tasks, where such a verifier is unavailable, we experiment with the following methods:

\begin{itemize}[leftmargin=16pt, topsep=1pt]
    \item \textbf{Confidence-based.} Following~\citet{kang2025scalablebestofnselectionlarge}, we adopt the \textit{self-certainty} metric as a lightweight, confidence-based selection method. Specifically, self-certainty quantifies model confidence by computing the KL divergence between the predicted token distribution and a uniform distribution. After computing the self-certainty scores ${c_i}_{i=1}^{N}$ for the set of predictions $\mathcal{P}$, we select the final answer using Borda voting, as described in the same work. For benchmarks where equivalence checking is not applicable (LiveCodeBench), we simply select the response with the highest self-certainty score.
    \item \textbf{Majority voting.} We report results based on consensus (majority vote) for tasks that have exact answers (e.g., math problem solving and science), following prior work~\citep{deepseek-r1, wang2023selfconsistencyimproveschainthought}. Specifically, given a set of answers $\{a_i\}_{i=1}^n$ extracted from all predictions in $\mathcal{P}$, we define consensus as $\text{cons@}n=\text{argmax}_a\sum_{i=1}^N\mathbb{1}(a_i=a)$. For cases where $k < N$, we estimate $\text{cons@}k$ by randomly sampling $k$ answers from the full set of $N$ predictions, computing the majority vote, and averaging accuracy over multiple trials using Monte Carlo simulation.
    
\end{itemize} 

\paragraph{Metrics}
We define \textbf{latency} as the maximum number of tokens generated averaged across the $N$ trials and across the dataset. This metric is practically important, as lower latency translates to faster response times for users, an essential goal in real-world applications.

\begin{figure}[t]
  \centering
  \includegraphics[width=\textwidth]{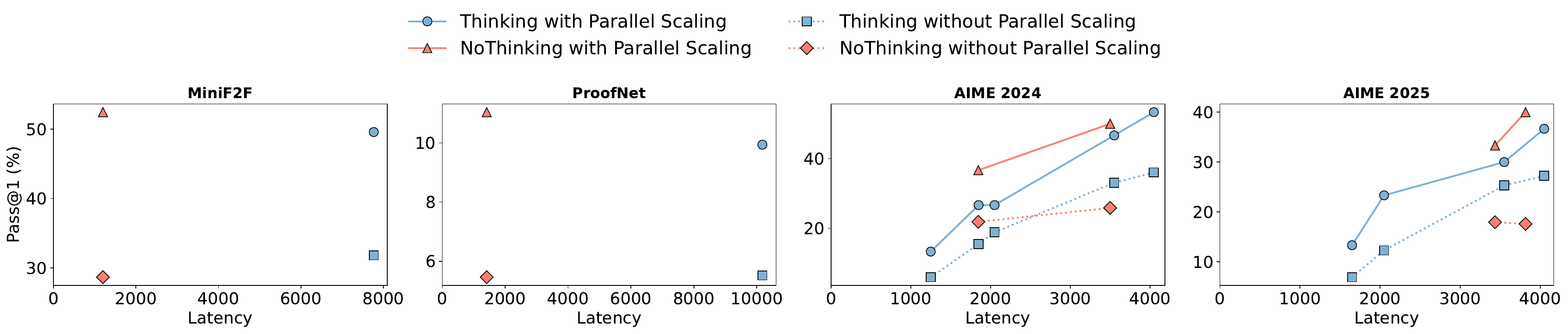}
    \includegraphics[width=0.75\textwidth]{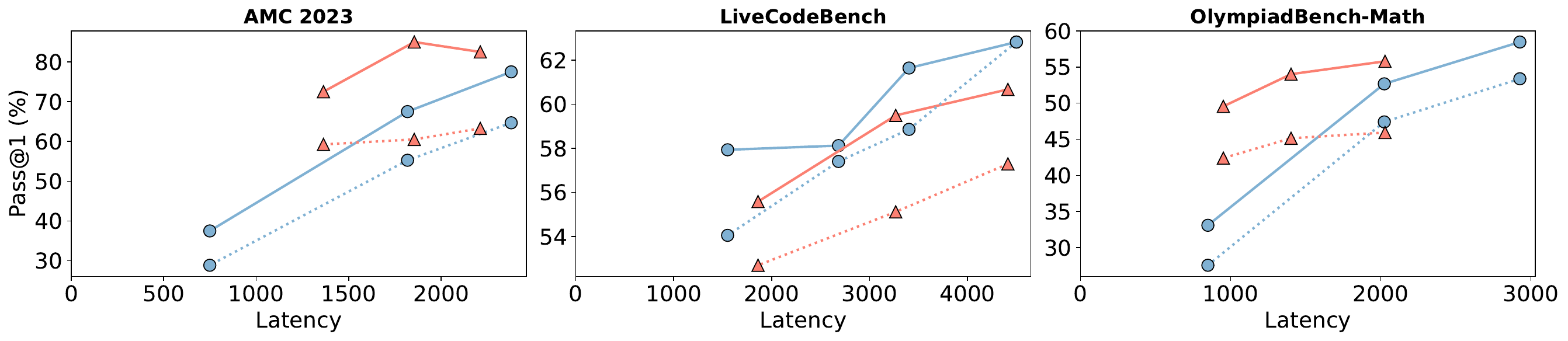}


  \caption{Pass@1 results across all benchmarks, comparing \nothinking and \thinking, with and without parallel scaling. 
  }
  \label{fig:best_of_n}
\end{figure}

\begin{table}[t]
  \centering
  \small
  \begin{adjustbox}{width=\textwidth}
    \begin{tabular}{ccrrP{1.5cm}P{1.5cm}P{1.5cm}}
    \toprule
    \multirow{2}[4]{*}{\textbf{Task}} & \multirow{2}[4]{*}{\textbf{Thinking}} & \multirow{2}[4]{*}{\textbf{BF (tokens)}} & \multirow{2}[4]{*}{\textbf{Pass@K}} & \multicolumn{3}{c}{\textbf{Selection Methods (Pass@1)}} \\
\cmidrule{5-7}          &       &       &       & \textbf{Majority Voting} & \textbf{Confidence + Highest} & \textbf{Confidence + Voting} \\
    \midrule
    \multirow{2}[4]{*}{AIME 2024} & \thinking & 3500  & 73.33 & 43.33 & 40.00 & \textbf{46.67} \\
\cmidrule{2-7}          & \nothinking & 3500  & 77.30 & 46.67 & 20.00 & \textbf{50.00} \\
    \midrule
    \multirow{2}[4]{*}{AIME 2025} & \thinking & 3500  & 40.00 & \textbf{30.00} & \textbf{30.00} & \textbf{30.00} \\
\cmidrule{2-7}          & \nothinking & 3500  & 53.73 & \textbf{33.33} & 20.00 & \textbf{33.33} \\
    \midrule
    \multirow{2}[4]{*}{AMC 2023} & \thinking & 2400  & 92.50 & \textbf{77.50} & 65.00 & 77.50 \\
\cmidrule{2-7}          & \nothinking & 2400  & 95.00 & 77.50 & 57.50 & \textbf{85.00} \\
    \bottomrule
    \end{tabular}%
  \end{adjustbox}
\caption{\textbf{Comparison of Best-of-$N$ selection methods—majority voting, confidence+highest, and confidence+voting—on selected experiments.} In ``confidence+highest'', we select the response with the highest model confidence. In ``confidence+voting'', responses are ranked by confidence from high to low and assigned scores using  $v(r) = (N - r + 1)^p$, where $r$ is the rank and $p = 0.5$; scores are then summed across responses that yield the same answer.  The best-performing method for each task is highlighted in bold. We also report pass@$k$ as an upper bound, where $k$ is the number of samples used in Best-of-$N$ selection.}
  \label{tab:best-of-n-method-comparison}%
\end{table}%

\subsection{Results}\label{subsec:best-of-n-results}
We present pass@1 results for \thinking and \nothinking across all benchmarks in~\autoref{fig:best_of_n}. The performance of a single sampled response is referred to as pass@1 without parallel scaling, while the accuracy of Best-of-$N$ selection over multiple samples is considered pass@1 with parallel scaling. For tasks without verifiers, we use confidence-based results in the figure and present ablation results in~\autoref{tab:best-of-n-method-comparison} for selected experiments. This table compares the Best-of-$N$ methods discussed in Section~\ref{subsec:parallel-v-sequential}. Confidence-based selection generally outperforms majority voting. We also report pass@$k$ accuracy as an upper bound on pass@1 when using parallel scaling, which is included in~\autoref{tab:best-of-n-method-comparison}. An extended version of the table is available in Appendix~\ref{subsec:best-of-n-methods-full}.

\paragraph{Perfect Verifiers}
\nothinking, when combined with parallel scaling, offers a highly efficient alternative to traditional sequential approaches—achieving similar or better accuracy with significantly lower latency and token usage. As shown in the first two plots of~\autoref{fig:best_of_n}, \nothinking achieves comparable or even higher performance than \thinking, while incurring much lower latency. Without parallel scaling, \nothinking closely matches \thinking in accuracy at a fraction of the latency. If a perfect verifier is available, selecting the best among $k$ sampled responses would achieve pass@$k$ accuracy. When combined with parallel scaling, \nothinking matches the accuracy of \thinking without budget forcing and parallel scaling—the representative sequential scaling baseline—while reducing latency by a factor of 7. Moreover, on both the MiniF2F and ProofNet datasets, \nothinking achieves the same accuracy using 4× fewer output tokens, highlighting its computational efficiency. These results underscore the effectiveness of parallel sampling in verifier-available settings. Additional details are provided in Appendix~\ref{subsec:verifier-results}.

\paragraph{Simple Best-of-$N$ Methods}
\nothinking, when paired with parallel scaling and confidence-based selection, consistently outperforms \thinking across most benchmarks under low token budgets. \autoref{fig:best_of_n} (last five plots) presents confidence-based selection results across several benchmarks, comparing \thinking and \nothinking under controlled token usage. We focus on this low-budget regime for two reasons: (1) it aligns with our primary interest in efficient inference, and (2) setting \texttt{max\_tokens} too high often results in excessively long and incoherent outputs (``babbling''), which inflates latency and diminishes the value of comparison. As expected, parallel scaling improves pass@1 performance for both \thinking and \nothinking. However, across all math benchmarks, \nothinking consistently dominates the Pareto frontier of \thinking with parallel scaling, demonstrating superior accuracy–budget tradeoffs. On AMC 2023 and OlympiadBench, \nothinking outperforms \thinking consistently, regardless of whether parallel scaling is used. Notably, even when compared to full \thinking (\thinking without budget forcing), \nothinking achieves a higher pass@1 score (55.79 vs. 54.1) while reducing latency by a factor of 9.

\nothinking is less effective on LiveCodeBench, which appears to be an outlier. This may be due to the limitations of confidence-based selection for coding tasks, where voting strategies cannot be applied without exact-match outputs. In these cases, we fall back to selecting the highest-confidence response, which is less reliable. As shown in~\autoref{tab:best-of-n-method-comparison}, this approach consistently underperforms (often by a large margin) compared to voting-based methods on tasks where voting is applicable.

Overall, these results highlight the effectiveness of \nothinking in verifier-free settings when combined with parallel sampling and strong selection strategies.

\paragraph{Summary} The surprising pass@$k$ performance of \nothinking as $k$ increases can be further leveraged through parallel scaling, enabling improved pass@1 results at similar—or even significantly lower—latency (up to 9× reduction). For tasks equipped with perfect verifiers, this approach also reduces total token usage by up to 4x with similar or better accuracy.
\section{Conclusion}

Large language models generating a lengthy ``thinking'' process before producing solutions have achieved strong results on reasoning tasks. In this paper, we question the necessity of this process by introducing \nothinking, a simple yet effective prompting approach that bypasses the explicit thinking process. We show that the same models, without long thinking chains, can perform comparably to or better than their \thinking counterparts in pass@$k$ as  k  increases, while using significantly fewer tokens. Under comparable token budgets, \nothinking consistently outperforms the traditional thinking results at most values of $k$. Furthermore, we show that \nothinking can be combined with best-of-$N$ selection methods to achieve better accuracy–latency tradeoffs than are possible with standard \thinking. 

We hope our research encourages a reconsideration of the necessity of lengthy thinking processes, while  providing a highly competitive reference for strong reasoning performance in low-budget settings and at low latency.

\section*{Acknowledgements}
This work was supported by the Sky Computing Lab at UC Berkeley. We are grateful to Databricks for providing compute resources and thank Jonathan Frankle for facilitating access. We also thank Kaylo Littlejohn (UC Berkeley) and Zhaoyu Li (University of Toronto) for their valuable feedback on the draft.

\newpage
\appendix
\section{Method Details}

\subsection{Budget Forcing Details}\label{app:budget-forcing-details}
To compare \thinking and \nothinking under similar token usage in an out-of-the-box manner, we adopt a technique from~\citet{simplescaling}, where a generation budget is enforced by setting the \texttt{max\_tokens} parameter during the initial model call. If the generation reaches this limit without producing a stop token (i.e., the reasoning is truncated), we prompt the model to produce a final answer directly by appending an \texttt{<|end\_of\_thinking|>} tag (for \thinking only) followed by a final answer prefix. For math and science tasks, we use the prompt string ``\texttt{Final Answer:}''. For coding tasks, we append code delimiters (i.e., \verb|```|) to ensure the model generates code directly. For a given budget, we first run \nothinking and record its average token usage, since its responses can vary in length and are not explicitly constrained. We then apply a budget that is slightly lower than the average token usage of \nothinking to \thinking because \thinking is more likely to be truncated under budget constraints. This setup helps ensure that the total output length of \thinking (including truncated reasoning and final answer) is roughly comparable to that of \nothinking.

\section{Additional Results}
\subsection{Results on Tasks with Verifiers}
\label{subsec:verifier-results}
\begin{table}[htbp]
  \centering
  \footnotesize
    \vspace{-1mm}
    \begin{tabular}{@{}lccccccr@{}}
    \toprule
    \multirow{2}[4]{*}{Task} & \multicolumn{3}{c}{Full Thinking (Pass@1)} & \multicolumn{4}{c}{NoThinking (Pass@$k$)} \\
\cmidrule{2-7}          & Latency $\downarrow$  & Total token $\downarrow$ & Accuracy & Latency $\downarrow$ & Total Token $\downarrow$ & Accuracy & $k$ \\
    \midrule
    MiniF2F & 7767  & 743956 & 34.84 & 1200 (-6.47x)  & 190349 (-3.91x) & 36.38 & 2 \\
    Proofnet & 10178 & 604507 & 5.52  & 1413 (-7.20x)  & 145921 (-4.14x) & 6.90 & 2 \\
    \bottomrule
    \end{tabular}%
    \caption{Costs for the \nothinking method to match full \thinking (pass@1) on tasks with verifiers. The “Total tokens” column reports the cumulative number of tokens used across all problems and samples to achieve the corresponding accuracy.}
    \label{tab:verifiers_results}%
\end{table}%
 
\subsection{More Models}
\label{sec:more_models}
\begin{figure}[htbp]
    \centering
    \includegraphics[width=0.8\linewidth]{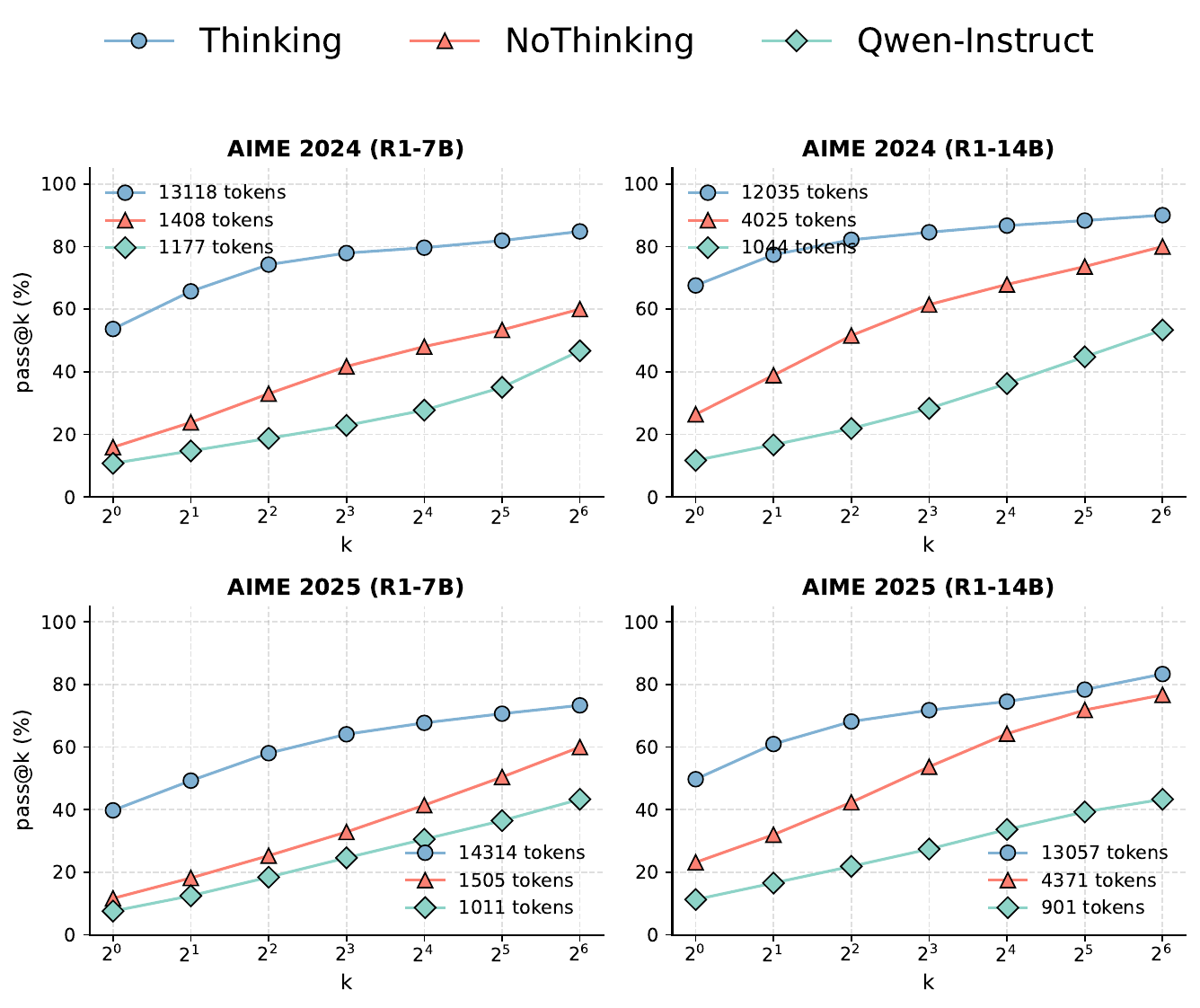}
    
    \caption{\thinking vs. \nothinking vs. Qwen-Instruct without token budget controlled on DeepSeek-R1-Distill-Qwen-7B and -14B models. Blue, orange, and green indicate \thinking, \nothinking, and Qwen-Instruct, respectively. We use a Qwen-Instruct model of the same size as the reasoning model for comparison.}
    \label{fig:7b_14b_no_bf}
\end{figure}
\begin{figure}[htbp]
    \centering
    \includegraphics[width=\linewidth]{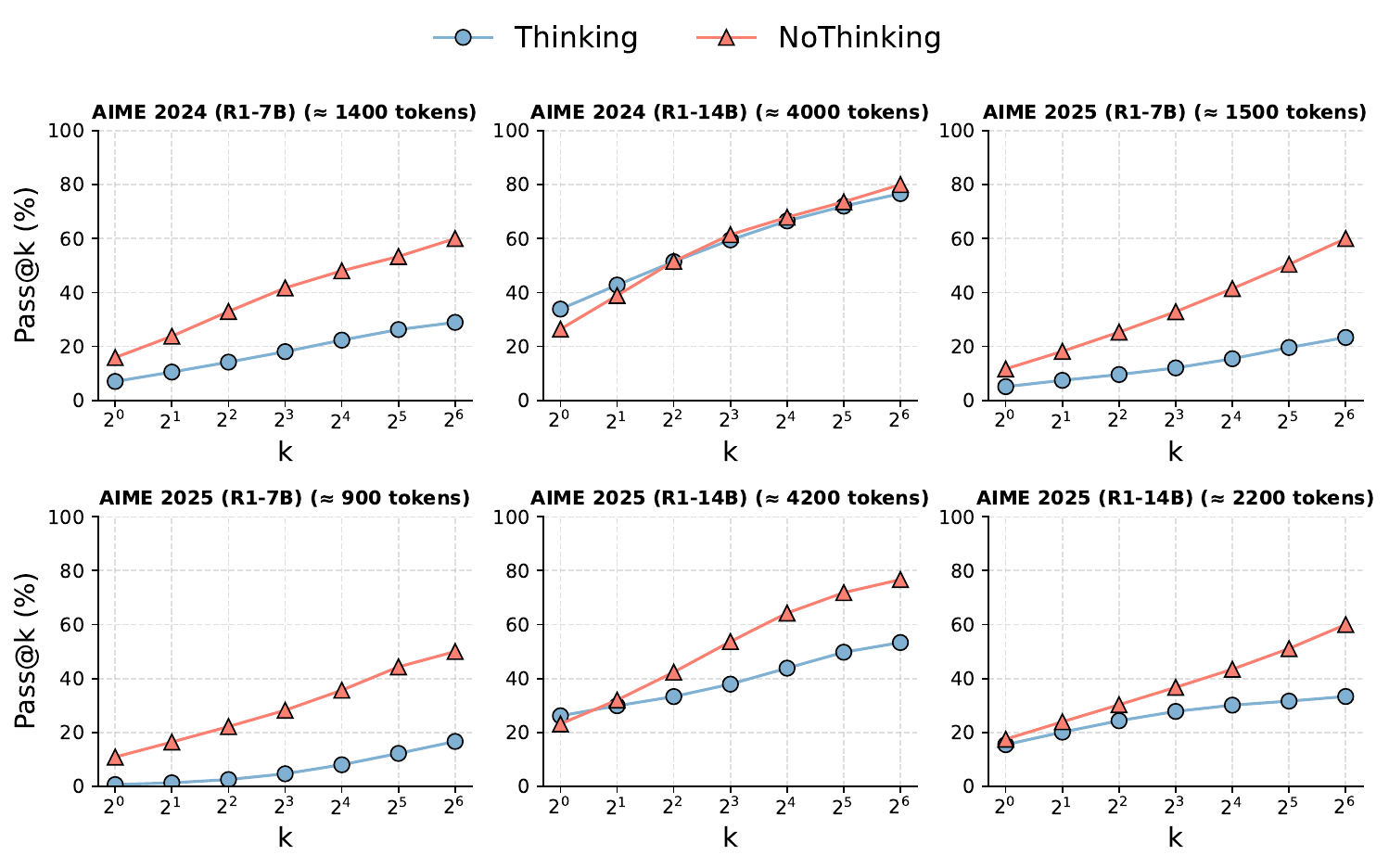}
    \caption{\thinking vs. \nothinking with token budget controlled on DeepSeek-R1-Distill-Qwen-7B and -14B models. Orange and blue indicate \nothinking and Thinking, respectively. Under comparable token budget, \nothinking consistently outperforms Thinking as $k$ increases.}
    \label{fig:7b_14b_bf}
\end{figure}
We experiment with smaller models (7B and 14B) from the R1-series on AIME tasks and observe similar behavior to that discussed in Section~\ref{sec:main-exp}. As shown in~\autoref{fig:7b_14b_no_bf}, notably, on the 14B model, \nothinking initially lags behind but rapidly catches up as $k$ increases, ultimately approaching the performance of \thinking. \nothinking on the 7B model already produces very few output tokens, so its performance does not increase as rapidly at $k=64$ compared to larger models. However, it still widens the gap with Qwen. We expect it to catch up to \thinking with larger values of $k$.

We apply budget forcing to Thinking to match the token usage of NoThinking (\autoref{fig:7b_14b_bf}). On the 7B model, \nothinking outperforms from the very beginning and continues to widen the gap. On the 14B model, it initially lags behind but eventually catches up. Exploring different budget settings would be an interesting direction for future work.

\subsection{Best-of-$N$ Methods Comparison}
\label{subsec:best-of-n-methods-full}
\autoref{tab:best-of-n-method-comparison-full} presents the results of various best-of-$N$ methods across all experiments. It is an extended version of~\autoref{tab:best-of-n-method-comparison}.
\begin{table}[t]
  \centering
  \small
  \begin{adjustbox}{width=\textwidth}
\begin{tabular}{ccrrP{1.5cm}P{1.5cm}P{1.5cm}}
\toprule
\multirow{2}[4]{*}{\textbf{Task}} & \multirow{2}[4]{*}{\textbf{Thinking}} & \multirow{2}[4]{*}{\textbf{BF (tokens)}} & \multirow{2}[4]{*}{\textbf{Pass@$k$}} & \multicolumn{3}{c}{\textbf{Selection Methods (Pass@1)}} \\
\cmidrule{5-7}          &       &       &       & \textbf{Majority Voting} & \textbf{Confidence + Highest} & \textbf{Confidence + Voting} \\
\midrule
\multirow{7}[14]{*}{\textbf{AIME 2024}} & \multirow{5}[10]{*}{\thinking} & 4000  & 80.00 & \textbf{53.33} & 33.33 & \textbf{53.33} \\
\cmidrule{3-7}          &       & 3500  & 73.33 & 43.33 & 40.00 & \textbf{46.67} \\
\cmidrule{3-7}          &       & 2000  & 50.00 & \textbf{26.67} & \textbf{26.67} & \textbf{26.67} \\
\cmidrule{3-7}          &       & 1800  & 50.00 & \textbf{26.67} & 13.33 & \textbf{26.67} \\
\cmidrule{3-7}          &       & 1200  & 23.33 & \textbf{13.33} & 3.33  & \textbf{13.33} \\
\cmidrule{2-7}          & \multirow{2}[4]{*}{\nothinking} & 3500  & 77.30 & 46.67 & 20.00 & \textbf{50.00} \\
\cmidrule{3-7}          &       & 1800  & 63.33 & \textbf{36.67} & 23.33 & \textbf{36.67} \\
\midrule
\multirow{6}[12]{*}{\textbf{AIME 2025}} & \multirow{4}[8]{*}{\thinking} & 4000  & 50.00 & 36.67 & 26.67 & \textbf{40.00} \\
\cmidrule{3-7}          &       & 3500  & 40.00 & \textbf{30.00} & \textbf{30.00} & \textbf{30.00} \\
\cmidrule{3-7}          &       & 2000  & 30.00 & \textbf{23.33} & 10.00 & \textbf{23.33} \\
\cmidrule{3-7}          &       & 1600  & 23.33 & \textbf{13.33} & 10.00 & \textbf{13.33} \\
\cmidrule{2-7}          & \multirow{2}[4]{*}{\nothinking} & 4000  & 57.70 & 36.67 & 23.33 & \textbf{40.00} \\
\cmidrule{3-7}          &       & 3500  & 53.73 & \textbf{33.33} & 20.00 & \textbf{33.33} \\
\midrule
\multirow{6}[12]{*}{\textbf{AMC 2023}} & \multirow{3}[6]{*}{\thinking} & 2400  & 92.50 & \textbf{77.50} & 65.00 & 77.50 \\
\cmidrule{3-7}          &       & 1800  & 90.00 & \textbf{67.50} & 62.50 & 65.00 \\
\cmidrule{3-7}          &       & 700   & 70.00 & \textbf{37.50} & 32.50 & \textbf{37.50} \\
\cmidrule{2-7}          & \multirow{3}[6]{*}{\nothinking} & 3000  & 95.00 & 80.00 & 62.50 & \textbf{82.50} \\
\cmidrule{3-7}          &       & 2400  & 95.00 & 77.50 & 57.50 & \textbf{85.00} \\
\cmidrule{3-7}          &       & 1600  & 95.00 & \textbf{72.50} & 57.50 & \textbf{72.50} \\
\midrule
\multirow{6}[12]{*}{\textbf{OlympiadBench}} & \multirow{3}[6]{*}{\thinking} & 3000  & 69.44 & \textbf{59.05} & 53.12 & 58.46 \\
\cmidrule{3-7}          &       & 2000  & 65.88 & 52.37 & 48.81 & \textbf{52.57} \\
\cmidrule{3-7}          &       & 800   & 43.77 & 32.64 & 28.19 & \textbf{33.09} \\
\cmidrule{2-7}          & \multirow{3}[6]{*}{\nothinking} & 3000  & 71.22 & 55.49 & 49.85 & \textbf{55.79} \\
\cmidrule{3-7}          &       & 1800  & 70.47 & 53.12 & 46.14 & \textbf{54.01} \\
\cmidrule{3-7}          &       & 1000  & 66.96 & \textbf{50.00} & 43.03 & 49.55 \\
\bottomrule
\end{tabular}%

    \end{adjustbox}
\caption{\textbf{Comparison of Best-of-$N$ selection methods—majority voting, confidence+highest, and confidence+voting—on benchmarks with exact-match answers (i.e., math benchmarks), evaluated under different token budgets for budget forcing (BF).} In ``confidence+highest'', we select the response with the highest model confidence. In ``confidence+voting'', responses are ranked by confidence from high to low and assigned scores using  $v(r) = (N - r + 1)^p$, where $r$ is the rank and $p = 0.5$; scores are then summed across responses that yield the same answer.  The best-performing method for each task is highlighted in bold. We also report pass@$k$ as an upper bound, where $k$ is the number of samples used in Best-of-$N$ selection.}
  \label{tab:best-of-n-method-comparison-full}%
\end{table}%

\section{Example Generations for Different Tasks}
\label{subsec:example-outputs}
We present three examples of complete model traces that illustrate math problem solving, coding, and formal theorem proving. Each example includes a comparison between \thinking and \nothinking without budget forcing.
\subsection{AIME 2024} 

Below, we present the full input and output traces for both \thinking and \nothinking for the example shown in~\autoref{fig:overview}.

\definecolor{thinking_color}{RGB}{194, 213, 247}

\begin{tcolorbox}[
  title=\thinking (6099 Completion Tokens),
  breakable,   
  fonttitle=\bfseries,
  enhanced,                        
  colback=thinking_color!10,           
  colbacktitle=thinking_color,         
  coltitle=black,                 
  colframe=thinking_color!80!black,    
  coltext=black,                  
  boxrule=0.5pt,
  arc=2mm
]
\small
\textbf{User:} Return your final response within \boxed{}. There exist real numbers $x$ and $y$, both greater than 1, such that $\log_x\left(y^x\right)=\log_y\left(x^{4y}\right)=10$. Find $xy$.
You should carefully think about the problem and reason step by step.
\\*[3mm]
\textbf{Assistant:} 
\\*[3mm]
\colorbox{thinking_color}{\textless think \textgreater}

Alright, so I have this logarithmic problem here, and I need to find the product $xy$ where $x$ and $y$ are real numbers greater than 1. The problem states that log base $x$ of $y^x$ equals 10, and log base $y$ of $x^{4y}$ also equals 10. Hmm, okay, let's try to break this down step by step.

First, let me recall the logarithm change of base formula because that seems relevant here. The formula is $\log_b(a) = (\ln a)/(\ln b)$, right? So, if I have a logarithm with some base, I can convert it to natural logarithms or any other base. That might help me manipulate the equations.

So, the first equation is $\log_x(y^x) = 10$. Let me rewrite that using the change of base formula. It becomes
\[
\frac{\ln(y^x)}{\ln x} = 10.
\]
Similarly, the second equation is $\log_y(x^{4y}) = 10$, which would be
\[
\frac{\ln(x^{4y})}{\ln y} = 10.
\]

Wait, maybe I can simplify these expressions before applying the change of base. Remember, $\log_b(a^c) = c \cdot \log_b(a)$. So, for the first equation, $\log_x(y^x)$ is equal to $x \cdot \log_x(y)$. Similarly, the second equation, $\log_y(x^{4y})$ is equal to $4y \cdot \log_y(x)$.

So, substituting back, the first equation becomes $x \cdot \log_x(y) = 10$, and the second equation becomes $4y \cdot \log_y(x) = 10$.

Hmm, now, do I remember any relationship between $\log_x(y)$ and $\log_y(x)$? Yes, they are reciprocals of each other, because $\log_x(y) = \frac{1}{\log_y(x)}$. That might come in handy here.

Let me denote $\log_x(y)$ as some variable to make it easier. Let's say $a = \log_x(y)$. Then, $\log_y(x)$ would be $1/a$ because of the reciprocal relationship.

So substituting back into the equations, the first equation becomes x * a = 10, and the second equation becomes $4y \cdot (1/a) = 10$. So now, I have two equations:
\begin{itemize}
    \item 1) $x \cdot a = 10$
    \item 2) $4y / a = 10$
\end{itemize}

Okay, so from the first equation, \( x = \frac{10}{a} \), and from the second equation, \( 4y = 10a \), so \( y = \frac{10a}{4} \), which simplifies to \( y = \frac{5a}{2} \).

So now, I have expressions for \( x \) and \( y \) in terms of \( a \). So, \( x = \frac{10}{a} \) and \( y = \frac{5a}{2} \).

Wait, but I also know that \( a = \log_x(y) \). Let me write that in terms of \( x \) and \( y \):

\[
\log_x(y) = a
\]

But since \( x = \frac{10}{a} \) and \( y = \frac{5a}{2} \), substituting back:

\[
\log_{\frac{10}{a}}\left(\frac{5a}{2}\right) = a
\]

Hmm, that seems a bit complicated, but maybe I can rewrite this using logarithm properties or exponentials.

Remember that \( \log_b(c) = d \) is equivalent to \( b^d = c \). So, in this case,

\[
\left(\frac{10}{a}\right)^a = \frac{5a}{2}
\]

So, $(10/a)^a = (5a)/2$.

Hmm, that's an equation in terms of $a$, which I can try to solve.

Let me write that down:

\[
\left(\frac{10}{a}\right)^a = \frac{5a}{2}
\]

This seems tricky because the variable is both in the base and exponent. Maybe taking the natural logarithm on both sides would help.

So, taking \( \ln \) of both sides:

\[
\ln\left[\left(\frac{10}{a}\right)^a\right] = \ln\left(\frac{5a}{2}\right)
\]

Using the logarithmic power rule on the left side:

\[
a \cdot \ln\left(\frac{10}{a}\right) = \ln(5) + \ln(a) - \ln(2)
\]

Let me expand that:

\[
a \cdot \ln(10) - a \cdot \ln(a) = \ln(5) + \ln(a) - \ln(2)
\]

Let me denote \( t = \ln(a) \), then \( a = e^t \). Let me try substituting that in.

So, equation becomes:

\[
e^t \cdot \ln(10) - e^t \cdot t = \ln(5) + t - \ln(2)
\]

Hmm, still seems messy, but maybe I can factor out \(e^t\) on the left side:

\[
e^t (\ln(10) - t) = \ln(5) + t - \ln(2)
\]

This is a transcendental equation, which might not have an algebraic solution. Maybe I need to solve it numerically or make an intelligent guess.

Alternatively, perhaps there's a substitution or some way to rearrange terms.

Wait, let's step back and see if there's another approach. Maybe instead of expressing $x$ and $y$ in terms of $a$, I can express both equations in terms of $a$ and then find a relationship between $x$ and $y$.

Wait, another thought: since I have $x = 10/a$ and $y = 5a/2$, maybe I can express $y$ in terms of $x$.

So, from $x = 10/a$, we get $a = 10/x$. Then, $y = 5a/2 = 5*(10/x)/2 = 50/(2x) = 25/x$.

So, $y = 25/x$.

Therefore, the product $xy = x \cdot (25/x) = 25$. Wait, so is it that simple?

Wait, hold on, if $y = 25/x$, then $xy = 25$. But before I conclude that, I need to make sure that this is consistent with the original equations.

Because sometimes when you manipulate equations, you can introduce extraneous solutions or lose constraints.

So, let me verify if this holds.

Given that xy = 25, so y = 25/x.

Then, plugging back into the original equations:

First equation: $\log_x(y^x) = 10$.

Which is $\log_x((25/x)^x) = 10$.

Simplify inside the log: $(25/x)^x = 25^x / x^x$.

So, $\log_x(25^x / x^x) = 10$.

Using logarithm rules, $\log_x(25^x) - \log_x(x^x) = 10$.

Which is $x \cdot \log_x(25) - x \cdot \log_x(x) = 10$.

Simplify: $x \cdot \log_x(25) - x \cdot 1 = 10$.

So, $x \cdot \log_x(25) = x + 10$.

Similarly, $\log_x(25)$ can be written as $\ln(25)/\ln(x)$.

So, $x\cdot(\ln(25)/\ln(x)) = x + 10$.

Hmm, that's the same as $(x / \ln(x)) \cdot \ln(25) = x + 10$.

Not sure if that helps. Let me see if I can manipulate the other equation.

Second equation: $\log_y(x^{4y}) = 10$.

Given that $y = 25/x$, so $\log_{25/x}(x^{4\cdot(25/x)}) = 10$.

Simplify exponent: $4\cdot(25/x) = 100/x$.

So, $\log_{25/x}(x^{100/x}) = 10$.

Again, using logarithm power rule: $(100/x) * \log_{25/x}(x) = 10$.

So, $(100/x) * [\ln(x)/\ln(25/x)] = 10$.

Simplify denominator: $\ln(25/x) = \ln(25) - \ln(x)$.

So, $(100/x) * [\ln(x)/(\ln(25) - \ln(x))] = 10$.

Multiply both sides by $x$:

\[100 \cdot [\ln(x)/(\ln(25) - \ln(x))] = 10x\]

Divide both sides by 10:

\[10 \cdot [\ln(x)/(\ln(25) - \ln(x))] = x\]

Hmm, so $10 \cdot ln(x) = x \cdot (\ln(25) - \ln(x))$.

Bring all terms to one side:

\[10 \ln(x) + x \ln(x) - x \ln(25) = 0\]

Factor terms:

\[\ln(x) \cdot (10 + x) - x \ln(25) = 0\]

Hmm, this still seems complicated. Maybe I need to test if $x=5$, because 25 is 5 squared, so maybe $x=5$ is a solution.

Let me test $x=5$.

If $x=5$, then $y=25/5=5$.

So, $x=5$ and $y=5$.

Let me check the original equations:

First equation: $\log_5(5^5) = \log_5(3125) = 5$, but it's supposed to be $10$. Hmm, that doesn't work.

Wait, so $x=5$ and $y=5$ gives us $\log_x(y^x)=5$, not $10$. So, that's not a solution.

Hmm, maybe $x=$ something else.

Wait, maybe $x=$ something else. Let me think.

Alternatively, perhaps I made a wrong assumption when I set $y=25/x$. Let's see, because from the earlier step, I had $x=10/a$ and $y=5a/2$, which leads to $y=25/x$, but that may or may not satisfy the original logarithmic equations.

Alternatively, maybe I can use the first approach where I had $(10/a)^a = (5a)/2$.

Let me try to solve this equation numerically.

So, let me define $f(a) = (10/a)^a - (5a)/2$.

We can try plugging in some values for a to see where $f(a)=0$.

First, let me note that a must be positive because it's a logarithm.

Let me try $a=2$.

$f(2) = (10/2)^2 - (5*2)/2 = 5^2 - 10/2 = 25 - 5 = 20 >0$.

$a=3$: $f(3)=(10/3)^3 - (15)/2 \approx (3.333)^3 -7.5 \approx 37.037 -7.5\approx 29.537>0$.

$a=4$: $(10/4)^4 - (20)/2= (2.5)^4 -10\approx 39.0625 -10\approx 29.0625>0$.

$a=5$: $(10/5)^5 -25/2=2^5 -12.5=32 -12.5=19.5>0$.

$a=1$: $f(1)=10^1 -5/2=10 -2.5=7.5>0$.

$a=0.5$: $(10/0.5)^0.5 - (5*0.5)/2=20^0.5 -2.5\approx4.472 -2.5\approx 1.972>0$.

$a=0.25$: $(10/0.25)^0.25 - (5*0.25)/2=40^0.25 -0.625\approx2.514 -0.625\approx 1.889>0$.

Hmm, all positive. So $f(a)$ is positive at all these points. Wait, that suggests that $(10/a)^a$ is always greater than $(5a)/2$ for $a>0$? But that can't be, because when a approaches infinity, $10/a$ approaches $0$, so $(10/a)^a$ approaches $0$, and $(5a)/2$ approaches infinity. So, at some point, $f(a)$ must become negative.

Wait, but in the $a$ approaches $0$, $10/a$ approaches infinity, so $(10/a)^a$ approaches infinity$^0$, which is an indeterminate form, but often tends to $1$. For example, (something large)$^0=1$, but (something large)$^{\text{something small}}$ can be $e^{\ln(\text{something large}) \cdot \text{something small}}$ which could go to $1$ or something else.

Wait, but when a approaches $0+$, $(10/a)^a = e^{a \ln(10/a)} = e^{a (\ln10 - \ln a)}$.

As a approaches $0+$, $a (\ln 10 - \ln a)$ approaches $0*(\ln10 - (-infty))$ which is $0\cdot(-infty)$, which is undefined, but the limit might be finite.

Wait, actually, the limit as a approaches $0+$ of a $ln(1/a)$ is 0, because $ln(1/a)$ goes to infinity, but a goes to zero, and the product goes to zero.

So, $(10/a)^a$ approaches $e^0=1$ as a approaches $0+$.

So, $f(a)$ approaches $1 - 0=1>0$.

Similarly, as a approaches infinity, $(10/a)^a$ approaches $0$, and $(5a)/2$ approaches infinity, so $f(a)$ approaches negative infinity.

Therefore, by the Intermediate Value Theorem, there must be some $a >0$ where $f(a)=0$.

So, let me try $a=6$: $(10/6)^6 - (5*6)/2\approx (1.6667)^6 -15\approx 17.79 -15\approx 2.79>0$.

$a=7$: $(10/7)^7\approx(1.4286)^7\approx$? Let me compute step by step:

$1.4286^2\approx 2.0408$

$1.4286^4\approx (2.0408)^2a\approx 4.164$

$1.4286^6\approx (4.164)\cdot(2.0408)\approx 8.490$

$1.4286^7\approx 8.490\cdot 1.4286\approx 12.11$

So, $(10/7)^7\approx 12.11$, and $(5\cdot 7)/2=17.5$, so $f(7)=12.11 -17.5\approx -5.39<0$.

So, between $a=6$ and $a=7$, $f(a)$ crosses zero.

Let me try $a=6.5$:

\[
\left(\frac{10}{6.5}\right)^{6.5} - \frac{5 \cdot 6.5}{2} \approx (1.5385)^{6.5} - 16.25
\]

Compute \(1.5385^{6.5}\):

First, \(\ln(1.5385) \approx 0.433\)

Multiply by 6.5: \(0.433 \cdot 6.5 \approx 2.8145\)

Exponentiate: \(e^{2.8145} \approx 16.65\)

So, \(\left(\frac{10}{6.5}\right)^{6.5} \approx 16.65\)

Then, subtract 16.25: \(f(6.5) = 16.65 - 16.25 \approx 0.4 > 0\)

So, \(f(6.5) = 0.4 > 0\)

Next, try \(a = 6.6\):

\[
\left(\frac{10}{6.6}\right)^{6.6} - \frac{5 \cdot 6.6}{2} \approx (1.5152)^{6.6} - 16.5
\]

Compute \(\ln(1.5152) \approx 0.416\)

Multiply by 6.6: \(0.416 \cdot 6.6 \approx 2.7456\)

Exponentiate: \(e^{2.7456} \approx 15.63\)

So, \(\left(\frac{10}{6.6}\right)^{6.6} \approx 15.63\)

Subtract 16.5: \(f(6.6) = 15.63 - 16.5 \approx -0.87 < 0\)

So, \(f(6.6) \approx -0.87 < 0\)

So, between \(a = 6.5\) and \(a = 6.6\), \(f(a)\) crosses zero.

Try \(a = 6.55\):

\[
\left(\frac{10}{6.55}\right)^{6.55} - \frac{5 \cdot 6.55}{2} \approx (1.5267)^{6.55} - 16.375
\]

Compute \(\ln(1.5267) \approx 0.423\)

Multiply by 6.55: \(0.423 \cdot 6.55 \approx 2.771\)

Exponentiate: \(e^{2.771} \approx 16.0\)

So, \(\left(\frac{10}{6.55}\right)^{6.55} \approx 16.0\)

Subtract 16.375: \(f(6.55) = 16.0 - 16.375 \approx -0.375 < 0\)

Earlier, at \(a = 6.5\), \(f(a) = 0.4 > 0\)

Try linear approximation between \(a = 6.5\) and \(a = 6.6\):

- Change in \(a = 0.1\)
- Change in \(f(a) = -1.27\)
- Solve \(0.4 + (-1.27)(\Delta a / 0.1) = 0\)

\[
0.4 - 12.7 \cdot \Delta a = 0 \Rightarrow \Delta a = \frac{0.4}{12.7} \approx 0.0315
\]

So, \(a \approx 6.5 + 0.0315 \approx 6.5315\)

Try \(a = 6.53\):

\[
\left(\frac{10}{6.53}\right)^{6.53} - \frac{5 \cdot 6.53}{2} \approx (1.531)^{6.53} - 16.325
\]

\(\ln(1.531) \approx 0.427\)

\(0.427 \cdot 6.53 \approx 2.788\)

\(e^{2.788} \approx 16.25\)

So, \(f(6.53) = 16.25 - 16.325 \approx -0.075\)

Try \(a = 6.52\):

\[
\left(\frac{10}{6.52}\right)^{6.52} - \frac{5 \cdot 6.52}{2} \approx (1.534)^{6.52} - 16.3
\]

\(\ln(1.534) \approx 0.429\)

\(0.429 \cdot 6.52 \approx 2.796\)

\(e^{2.796} \approx 16.35\)

\(f(6.52) = 16.35 - 16.3 = 0.05 > 0\)

So, between \(a = 6.52\) and \(a = 6.53\), \(f(a)\) crosses zero.

Use linear interpolation:

\[
f(a) = f_1 + \frac{f_2 - f_1}{a_2 - a_1}(a - a_1)
\]

\[
0 = 0.05 + \frac{-0.075 - 0.05}{0.01}(a - 6.52)
\Rightarrow -0.05 = -12.5(a - 6.52)
\Rightarrow a = \frac{81.55}{12.5} \approx 6.524
\]

Thus, \(a \approx 6.524\), and then:

\[
x = \frac{10}{a} \approx \frac{10}{6.524} \approx 1.533, \quad
y = \frac{5a}{2} \approx \frac{5 \cdot 6.524}{2} \approx 16.31
\]

Then:

\[
xy \approx 1.533 \cdot 16.31 \approx 25
\]

Which confirms that the product is approximately 25.

Finally:

\[
e^u \cdot \frac{v}{u} = 10, \quad
4e^v \cdot \frac{u}{v} = 10
\]

Multiply:

\[
\left( \frac{v}{u} e^u \right) \cdot \left( \frac{u}{v} \cdot 4 e^v \right) = 100
\Rightarrow 4e^{u+v} = 100 \Rightarrow e^{u+v} = 25
\]

Take natural log: u + v=ln(25).

But u=ln x, v=ln y, so ln x + ln y=ln(25).

Which is ln(xy)=ln(25), so xy=25.

Oh! So, despite the complicated a, the product xy is exactly 25.

So, the answer is 25.

Therefore, I can conclude that xy=25.

\textbf{Final Answer:}
The value of \( xy \) is \boxed{25}.

\colorbox{thinking_color}{\textless /think \textgreater}

Given the equations \(\log_x(y^x) = 10\) and \(\log_y(x^{4y}) = 10\), we need to find the product \(xy\).

First, we rewrite the logarithmic equations using the change of base formula and properties of logarithms:

1. \(\log_x(y^x) = 10\) can be rewritten as \(x \log_x(y) = 10\). Let \(a = \log_x(y)\). Then, this equation becomes \(x \cdot a = 10\).
2. \(\log_y(x^{4y}) = 10\) can be rewritten as \(4y \log_y(x) = 10\). Since \(\log_y(x) = \frac{1}{\log_x(y)} = \frac{1}{a}\), this equation becomes \(4y \cdot \frac{1}{a} = 10\).

From these, we derive:
- \(x = \frac{10}{a}\)
- \(y = \frac{10a}{4} = \frac{5a}{2}\)

We then express \(y\) in terms of \(x\):
- Since \(x = \frac{10}{a}\), we have \(a = \frac{10}{x}\).
- Substituting \(a\) into the expression for \(y\), we get \(y = \frac{5a}{2} = \frac{5 \cdot \frac{10}{x}}{2} = \frac{50}{2x} = \frac{25}{x}\).

Thus, \(y = \frac{25}{x}\), which implies \(xy = x \cdot \frac{25}{x} = 25\).

To verify, we use natural logarithms and properties of logarithms:
- Let \(u = \ln x\) and \(v = \ln y\). The equations transform into:
  1. \(e^u \cdot \frac{v}{u} = 10\)
  2. \(4e^v \cdot \frac{u}{v} = 10\)

Multiplying these equations:
- \(\left(\frac{v}{u} e^u\right) \left(\frac{u}{v} 4 e^v\right) = 100\)
- Simplifying, we get \(4 e^{u+v} = 100\), leading to \(e^{u+v} = 25\).
- Taking the natural logarithm, we find \(u + v = \ln(25)\), which means \(\ln(x) + \ln(y) = \ln(25)\), or \(\ln(xy) = \ln(25)\). Therefore, \(xy = 25\).

Thus, the value of \(xy\) is \(\boxed{25}\).

\end{tcolorbox}

\definecolor{no_thinking_color}{RGB}{252, 225, 198}

\begin{tcolorbox}[
  title=\nothinking (373 Completion Tokens),
  fonttitle=\bfseries,
  breakable,                     
  enhanced,                        
  colback=no_thinking_color!10,           
  colbacktitle=no_thinking_color,         
  coltitle=black,                 
  colframe=no_thinking_color!80!black,    
  coltext=black,                  
  boxrule=0.5pt,
  arc=2mm
]
\small
\textbf{User:} Return your final response within \boxed{}. There exist real numbers $x$ and $y$, both greater than 1, such that $\log_x\left(y^x\right)=\log_y\left(x^{4y}\right)=10$. Find $xy$.
Please write the answer for this math problem directly without any thinking process.
\\*[3mm]
\textbf{Assistant:} 
\\*[3mm]
\colorbox{no_thinking_color}{\textless think \textgreater}

Okay I have finished thinking.

\colorbox{no_thinking_color}{\textless /think \textgreater}

Let's solve the problem.

We are given that \( \log_x(y^x) = 10 \) and \( \log_y(x^{4y}) = 10 \). We need to find \( xy \).

First, recall the logarithm power rule: \( \log_b(a^n) = n \cdot \log_b(a) \).

Applying this rule to the first equation:
\[
\log_x(y^x) = x \cdot \log_x(y) = 10
\]
This implies:
\[
\log_x(y) = \frac{10}{x}
\]
Let’s denote this as equation (1).

Now, apply the power rule to the second equation:
\[
\log_y(x^{4y}) = 4y \cdot \log_y(x) = 10
\]
This implies:
\[
\log_y(x) = \frac{10}{4y} = \frac{5}{2y}
\]
Let’s denote this as equation (2).

Recall that \( \log_b(a) = \frac{1}{\log_a(b)} \). Using this property, we can relate equations (1) and (2):
\[
\log_x(y) = \frac{1}{\log_y(x)} \implies \frac{10}{x} = \frac{2y}{5}
\]
Simplifying:
\[
\frac{10}{x} = \frac{2y}{5} \implies 10 \cdot 5 = 2y \cdot x \implies 50 = 2xy \implies xy = 25
\]
Thus, the value of \( xy \) is \( 25 \).

\boxed{25}
\end{tcolorbox}

\subsection{LiveCodeBench}
\begin{tcolorbox}[
  title=\thinking (3326 Completion Tokens),
  breakable,   
  fonttitle=\bfseries,
  enhanced,                        
  colback=thinking_color!10,           
  colbacktitle=thinking_color,         
  coltitle=black,                 
  colframe=thinking_color!80!black,    
  coltext=black,                  
  boxrule=0.5pt,
  arc=2mm
]
\small
\textbf{User:} Generate an executable Python function generated from the given prompt. The function should take stdin as input and print the output. Simply call the function after the definition. There are three cards with letters $\texttt{a}$, $\texttt{b}$, $\texttt{c}$ placed in a row in some order. You can do the following operation at most once: 
\begin{itemize}[leftmargin=5mm]
    \item Pick two cards, and swap them.  Is it possible that the row becomes $\texttt{abc}$ after the operation? Output "YES" if it is possible, and "NO" otherwise.
\end{itemize}

\textbf{Input}

The first line contains a single integer $t$ ($1 \leq t \leq 6$) — the number of test cases.

The only line of each test case contains a single string consisting of each of the three characters $\texttt{a}$, $\texttt{b}$, and $\texttt{c}$ exactly once, representing the cards.

\textbf{Output}

For each test case, output "YES" if you can make the row $\texttt{abc}$ with at most one operation, or "NO" otherwise.

You can output the answer in any case (for example, the strings "yEs", "yes", "Yes" and "YES" will be recognized as a positive answer).Sample Input 1:
6

abc

acb

bac

bca

cab

cba

\textbf{Sample Output 1:}

YES
YES
YES
NO
NO
YES

\textbf{Note}

In the first test case, we don't need to do any operations, since the row is already $\texttt{abc}$.

In the second test case, we can swap $\texttt{c}$ and $\texttt{b}$: $\texttt{acb} \to \texttt{abc}$.

In the third test case, we can swap $\texttt{b}$ and $\texttt{a}$: $\texttt{bac} \to \texttt{abc}$.

In the fourth test case, it is impossible to make $\texttt{abc}$ using at most one operation.
You should carefully think about the problem and reason step by step.
\\*[3mm]
\textbf{Assistant:} 
\\*[3mm]
\colorbox{thinking_color}{\textless think \textgreater}

Okay, I have to solve this programming problem. Let's read the problem carefully.

So, the task is: Given a string of three distinct characters (a, b, c), determine if it's possible to turn it into "abc" by performing at most one swap of two characters. If it can be done, output YES; otherwise, NO.

Hmm, right. So, each test case is a permutation of a, b, c. We need to check if either:
\begin{enumerate}[leftmargin=5mm]
    \item The string is already "abc" — then we can do nothing and it's a YES.
OR
    \item There exists exactly one swap that can turn it into "abc".
\end{enumerate}

Wait, but how do I determine whether a single swap can achieve this?

Let me think about it. For each input string, I can consider all possible single swaps and see if any of them result in "abc".

But wait, the string is only three characters long. So, the number of possible swaps is limited.

Each swap can be between any two positions. For example, for a string s of length 3, the possible swaps are (0,1), (0,2), (1,2). So three possible swaps.

So, for each test case, I can do the following:
\begin{itemize}[leftmargin=5mm]
    \item If the string is already "abc", output YES.
    \item Else, check each possible pair of indices, swap them, and see if the result is "abc". If any of the swaps result in "abc", output YES. Otherwise, NO.
\end{itemize}

That sounds manageable.

But wait, what about the case where the string is just one swap away from "abc"? Like, for example, the input is "acb". Swapping positions 1 and 2 gives "abc", so YES.

Another example: input is "bca". Let's see. The desired output is "abc". So let's see what swaps can do:

Original: b c a

Swap 0 and 1: c b a → not "abc".

Swap 0 and 2: a c b → not "abc".

Swap 1 and 2: b a c → still not "abc".

So in that case, it's a NO.

Wait, but the sample input 4 is "bca" and the output is NO. So that's correct.

Another example: input is "cba". Let's see:

Swap 0 and 2: a b c → yes. So output is YES.

So, the approach is:

For each test case s:
\begin{itemize}[leftmargin=5mm]
    \item If s is "abc", output YES.
    \item Else, for each possible pair of indices (i, j), i < j, create a new string where s[i] and s[j] are swapped. Check if any of these new strings equals "abc". If yes, output YES; else, NO.
\end{itemize}

Alternatively, since the string is small, I can also compute how many characters are in the correct position. For example, the desired string is "abc". So:

If the number of correctly placed characters is 2 or 3, then it's possible with one swap or zero.

Wait, let's think: If the string is "acb", then two characters are correct (a and c are in correct positions). Swapping the wrong ones (b and c) will fix it.

But wait, in "acb", a is in position 0 (correct), c is in position 1 (wrong), b is in position 2 (correct). So two correct. So swapping positions 1 and 2 (c and b) will give "abc".

So, another approach: count the number of mismatches between the input string and "abc".

If the number of mismatches is 0: output YES.

If the number is 2: then swapping those two will fix it. So output YES.

If the number is 1 or 3: then it's impossible with one swap.

Wait, that's a clever insight. Because:
\begin{itemize}[leftmargin=5mm]
    \item 0 mismatches: already correct.
    \item 1 mismatch: you can't fix it with one swap, because a swap affects two positions. If only one character is wrong, you can't swap to fix it without messing up another.
    \item 2 mismatches: you can swap the two wrong positions, which will fix both.
    \item 3 mismatches: for example, "cba" is "c", "b", "a". So all positions are wrong. How? Wait, "cba" has a in position 2, which is correct for "abc"? No, "abc" has a in 0, b in 1, c in 2. So "cba" has a in 2, which is wrong. So all three positions are wrong. So that's three mismatches. But wait, can a single swap fix this? Let's see.
\end{itemize}

Wait, "cba" → swapping 0 and 2 gives "a b c" → yes. So in this case, with three mismatches, a single swap can fix the problem. So, the earlier approach is incorrect.

Wait, so the number of mismatches isn't a sufficient condition. Because in the case of three mismatches, sometimes a single swap can fix it.

So, maybe the initial approach of checking all possible swaps is better.

Alternatively, perhaps the number of mismatches is 0, 2, or 3. Wait, because in the case of three mismatches, swapping the first and last characters can fix it.

Hmm, so perhaps the condition is: if the number of mismatches is 0, or if the number of mismatches is 2, or if the number of mismatches is 3 but the string is the reverse of "abc".

Wait, but "cba" is the reverse. So, for example, in "cba", swapping 0 and 2 gives "abc". So in this case, the number of mismatches is 3, but it's fixable with one swap.

So, this suggests that the number of mismatches is not the only factor. So, perhaps the initial approach of checking all possible swaps is more reliable.

So, for each test case, I can:
\begin{enumerate}[leftmargin=5mm]
    \item Check if the string is already "abc". If yes, output YES.

    Else:
    \item For each possible pair of indices (i, j) where i < j, swap the characters at these positions and see if the result is "abc".
    \item If any of these swaps result in "abc", output YES.
    \item Else, output NO.
\end{enumerate}

Since the string is only three characters, there are only three possible swaps to check, so this approach is feasible.

So, now, how to implement this in Python.

Let's outline the steps:

Read t test cases.

For each test case:

s = input string.

if s == 'abc':
    print YES
else:
    for each possible pair (i, j) in (0,1), (0,2), (1,2):
        create a list of the characters.
        swap the i-th and j-th elements.
        check if the new string equals 'abc'.
        if any do, set a flag to True and break.
    if any swap worked, print YES, else NO.

Wait, but wait a second. For a string of 3 characters, the possible swap pairs are (0,1), (0,2), (1,2), which are all possible unique pairs. So for each test case, we can generate all possible swaps.

But how to generate all possible swaps?

In code:
\begin{verbatim}
for i in range(3):
    for j in range(i+1, 3):
        # swap i and j
        lst = list(s)
        lst[i], lst[j] = lst[j], lst[i]
        new_s = ''.join(lst)
        if new_s == 'abc':
            # success
            flag = True
            break
    if flag:
        break
\end{verbatim}

If any of the swaps result in 'abc', then output YES.

Yes, that should work.

Now, let's test this logic against the sample input.

Sample Input 1:

6

abc → yes

acb → swap 1 and 2 → abc → yes.

bac → swap 0 and 1 → abc → yes.

bca → possible swaps:

swap 0 and 1: bac → bca? No. Wait, let's see. Original is bca. Swapping 0 and 1: cba → not abc. Swapping 0 and 2: a c b → acb → not abc. Swapping 1 and 2: b a c → bac → no. So none of the swaps give abc. So output is NO.

cab → swap 0 and 2 → abc → yes? Let's see: original is cab. Swap 0 and 2: gives abc. So yes.

cba → swap 0 and 2: a b c → yes.

Wait, but in the sample output, the last test case is cba, which outputs YES.

Wait, the sample input is 6 test cases:

Test case 1: abc → YES

Test case 2: acb → YES

Test case3: bac → YES

Test case4: bca → output is NO

Test case5: cab → YES

Test case6: cba → output is YES.

So, according to the sample output, the fifth test case is cab, which swapping 0 and 2 gives abc. So yes.

But according to the code, for cab, the code would check all swaps:

swap 0 and 1: 'cba' → not 'abc'

swap 0 and 2: 'abc' → yes.

So the code would output YES.

Wait, but according to the sample output, the fifth test case's output is NO. Wait no, let me recheck the sample input and output.

Wait the sample input is:

6

abc

acb

bac

bca

cab

cba

Sample Output:

YES

YES

YES

NO

NO

YES

Wait, so test case 5 is 'cab' → output is NO. But according to the code, swapping 0 and 2 in 'cab' gives 'abc', so why the sample output is NO?

Wait, that's conflicting. Wait, what's 'cab'?

Wait, 'cab' is c a b.

If I swap 0 and 2: positions 0 is c, 2 is b. So after swap, it's b a c → bac → which is not abc.

Wait wait, no. Wait 'cab' is 'c', 'a', 'b'. Swapping 0 and 2: 'b', 'a', 'c' → 'bac' which is not 'abc'.

Wait, but earlier I thought that swapping 0 and 2 in 'cab' would give 'abc', but that's incorrect.

Wait, maybe I'm making a mistake in the swap.

Wait 'cab' is index 0: 'c', 1: 'a', 2: 'b'.

If I swap 0 and 1: string becomes 'a c b' → 'acb' → not 'abc'.

If I swap 0 and 2: string becomes 'b a c' → 'bac' → not 'abc'.

If I swap 1 and 2: string becomes 'c b a' → 'cba' → not 'abc'.

So, none of the swaps for 'cab' results in 'abc'. So, the correct output is NO.

But according to my earlier approach, for 'cab' the code would generate all possible swaps and find none result in 'abc', so output NO.

But in my initial analysis, I thought that swapping 0 and 2 would fix it. Oh no, I made a mistake there.

So, the code is correct.

Wait, so how come the sample output for the fifth test case is NO?

Because for 'cab', we need to find if any swap can turn it into 'abc'.

But in 'cab', the correct positions are a is in position 1 (should be 0), b is in position 2 (should be 1), c is in position 0 (should be 2). So, the number of mismatches is 3.

But a single swap can't fix all three. Because a swap affects two positions. So, it's impossible.

So, the code will correctly output NO for 'cab'.

So, the code seems correct.

Another test case: 'cba'.

Swap 0 and 2: 'a' 'b' 'c' → yes. So output is YES.

So, the code handles that.

So, the code approach is correct.

Now, implementing this in Python.

The code structure is as follows:

Read t.

For each test case:

s = input().strip()

if s == 'abc':
    print('YES')
else:
    possible = False
    for i in range(3):
        for j in range(i+1, 3):
            lst = list(s)
            lst[i], lst[j] = lst[j], lst[i]
            if ''.join(lst) == 'abc':
                possible = True
                break
        if possible:
            break
    print('YES' if possible else 'NO')

Yes.

Now, let's test this code against the sample inputs.

Sample Input:

6

abc → yes.

acb → swap 1 and 2: yes.

bac → swap 0 and 1: yes.

bca → no.

cab → no.

cba → swap 0 and 2: yes.

Which matches the sample output.

So, the code should be correct.

So, the final code is as written.

\colorbox{thinking_color}{\textless /think \textgreater}

To solve this problem, we need to determine if we can transform a given string of three distinct characters ('a', 'b', 'c') into the string "abc" by performing at most one swap of two characters. If it is possible, we output "YES"; otherwise, we output "NO".

\textbf{Approach}
\begin{enumerate}[leftmargin=5mm]
    \item \textbf{Check if the string is already "abc"}: If the input string is already "abc", we can immediately output "YES".
    \item \textbf{Check all possible single swaps}: For each input string that is not "abc", we generate all possible strings that can be formed by swapping exactly two characters. There are three possible swaps for a string of length 3:
    \begin{itemize}
        \item Swap the first and second characters.
        \item Swap the first and third characters.
        \item Swap the second and third characters.
    \end{itemize}
    \item \textbf{Evaluate each swap}: For each swap, we check if the resulting string is "abc". If any swap results in "abc", we output "YES". If none of the swaps result in "abc", we output "NO".
\end{enumerate}

\textbf{Solution Code}

\begin{verbatim}
```python
t = int(input())
for _ in range(t):
    s = input().strip()
    if s == 'abc':
        print("YES")
        continue
    possible = False
    for i in range(3):
        for j in range(i + 1, 3):
            lst = list(s)
            lst[i], lst[j] = lst[j], lst[i]
            if ''.join(lst) == 'abc':
                possible = True
                break
        if possible:
            break
    print("YES" if possible else "NO")
```
\end{verbatim}

\textbf{Explanation}
\begin{itemize}[leftmargin=5mm]
    \item \textbf{Reading Input}: We read the number of test cases and then process each string.
    \item \textbf{Immediate Check}: If the string is already "abc", we print "YES".
    \item \textbf{Swapping Characters}: For each string that is not "abc", we generate all possible strings by swapping each pair of characters. We check if any of these swapped strings equals "abc".
    \item \textbf{Output Result}: Based on whether any valid swap results in "abc", we print "YES" or "NO".
\end{itemize}

This approach ensures that we efficiently check all possible single swaps to determine if the transformation is possible, leveraging the small size of the input string to keep the solution simple and efficient.
\end{tcolorbox}

\begin{tcolorbox}[
  title=\nothinking (256 Completion Tokens),
  fonttitle=\bfseries,
  breakable,                     
  enhanced,                        
  colback=no_thinking_color!10,           
  colbacktitle=no_thinking_color,         
  coltitle=black,                 
  colframe=no_thinking_color!80!black,    
  coltext=black,                  
  boxrule=0.5pt,
  arc=2mm
]
\small
\textbf{User:} Generate an executable Python function generated from the given prompt. The function should take stdin as input and print the output. Simply call the function after the definition. There are three cards with letters $\texttt{a}$, $\texttt{b}$, $\texttt{c}$ placed in a row in some order. You can do the following operation at most once: 
\begin{itemize}[leftmargin=5mm]
    \item Pick two cards, and swap them.  Is it possible that the row becomes $\texttt{abc}$ after the operation? Output "YES" if it is possible, and "NO" otherwise.
\end{itemize}

\textbf{Input}

The first line contains a single integer $t$ ($1 \leq t \leq 6$) — the number of test cases.

The only line of each test case contains a single string consisting of each of the three characters $\texttt{a}$, $\texttt{b}$, and $\texttt{c}$ exactly once, representing the cards.

\textbf{Output}

For each test case, output "YES" if you can make the row $\texttt{abc}$ with at most one operation, or "NO" otherwise.

You can output the answer in any case (for example, the strings "yEs", "yes", "Yes" and "YES" will be recognized as a positive answer).Sample Input 1:
6

abc

acb

bac

bca

cab

cba

\textbf{Sample Output 1:}

YES
YES
YES
NO
NO
YES

\textbf{Note}

In the first test case, we don't need to do any operations, since the row is already $\texttt{abc}$.

In the second test case, we can swap $\texttt{c}$ and $\texttt{b}$: $\texttt{acb} \to \texttt{abc}$.

In the third test case, we can swap $\texttt{b}$ and $\texttt{a}$: $\texttt{bac} \to \texttt{abc}$.

In the fourth test case, it is impossible to make $\texttt{abc}$ using at most one operation.
Please solve the above problem without the thinking process.
\\*[3mm]
\textbf{Assistant:} 
\\*[3mm]
\colorbox{no_thinking_color}{\textless think \textgreater}

Okay, I have finished thinking.

\colorbox{no_thinking_color}{\textless /think \textgreater}

The task is to determine if each given permutation of the letters 'a', 'b', and 'c' can be transformed into "abc" with at most one swap. 

Here's how to approach it:
\begin{enumerate}[leftmargin=5mm]
    \item \textbf{Check if the string is already "abc"}: If yes, output "YES".
    \item \textbf{If not}, check how many positions each character is away from its target position in "abc":
    \begin{itemize}
        \item If exactly two characters are out of place, swapping them will make the string "abc".
        \item If more than two are out of place, it's impossible with one swap.
    \end{itemize}
    \item \textbf{Return "YES" if any of the above conditions are met, else "NO"}.
\end{enumerate}

Here's the Python code implementing this logic:
\begin{verbatim}
```python
def can_form_abc(s):
    if s == 'abc':
        return True
    target = 'abc'
    # Check if exactly two elements are different
    diff = sum(1 for a, b in zip(s, target) if a != b)
    return diff == 2

t = int(input())
for _ in range(t):
    s = input().strip()
    if can_form_abc(s):
        print("YES")
    else:
        print("NO")
```
\end{verbatim}
\end{tcolorbox}

\subsection{ProofNet}

\begin{tcolorbox}[
  title=\thinking (1913 Completion Tokens),
  breakable,   
  fonttitle=\bfseries,
  enhanced,                        
  colback=thinking_color!10,           
  colbacktitle=thinking_color,         
  coltitle=black,                 
  colframe=thinking_color!80!black,    
  coltext=black,                  
  boxrule=0.5pt,
  arc=2mm
]
\small
\textbf{User:}
\\*[3mm]
\textbf{Assistant:} 
\\*[3mm]
\colorbox{thinking_color}{\textless think \textgreater}

Alright, so I have this theorem to prove: if $G$ is a finite group and $a$ is an element of $G$, then there exists a positive integer $n$, depending on $a$, such that a raised to the power of $n$ equals the identity element $e$ in $G$. Okay, let's break this down.

First, I need to recall what a group is. A group is a set equipped with an operation that combines any two elements to form a third element, satisfying four conditions: closure, associativity, the existence of an identity element, and the existence of inverse elements. Since $G$ is finite, it has a finite number of elements.

Now, the theorem is saying that for any element $a$ in $G$, if I keep multiplying a by itself repeatedly, at some point I'll get back to the identity element $e$. That seems intuitive because the group is finite, so the powers of $a$ can't keep producing new elements forever; they have to eventually repeat or cycle back.

Let me think about an example. Suppose $G$ is the group of integers modulo 5 under addition. Wait, but addition is commutative, and this is a finite group. For any element $a \in G$, adding it to itself enough times will eventually give me 0, which is the identity element in this case. For instance, take $a = 2$ in $\mathbb{Z}/5\mathbb{Z}$. Then $2 + 2 = 4$, $2 + 2 + 2 = 6 \equiv 1 \pmod{5}$, and so on, until adding 2 five times gives $10 \equiv 0 \pmod{5}$. So, $n = 5$ works here.

But the theorem is about multiplicative notation, so let's think of a multiplicative group. Maybe the multiplicative group of complex roots of unity. For example, consider the group of 4th roots of unity: $\{1, i, -1, -i\}$. Take $a = i$. Then $i^1 = i$, $i^2 = -1$, $i^3 = -i$, $i^4 = 1$. So, $n = 4$ works here.

These examples suggest that such an n exists, but I need to prove it in general for any finite group.

The proof provided earlier mentions that because $G$ is finite, the set $\{a, a^2, a^3, \ldots\}$ must be finite. Wait, is that always true? If $G$ is finite, then the number of distinct powers of $a$ must be finite because there are only finitely many elements in $G$. So, the sequence $a, a^2, a^3, \ldots$ must eventually repeat.

The proof then says that there exist $i$ and $j$ with $i \ne j$ such that $a^i = a^j$. Without loss of generality, assume $i > j$. Then, multiplying both sides by $(a^j)^{-1}$, which is $a^{-j}$, gives $a^{i-j} = e$. So, $n = i - j$ is a positive integer such that $a^n = e$.

That seems straightforward, but let me make sure I understand why $i$ and $j$ must exist. Since the set $\{a, a^2, a^3, \ldots\}$ is finite, by the pigeonhole principle, there must be some repetition. That is, there must be distinct exponents $i$ and $j$ where $a^i = a^j$. This is because we have infinitely many exponents but only finitely many elements, so some elements must repeat.

Wait, but the exponents are infinite, while the group is finite. So, the mapping from exponents to group elements can't be injective; it must eventually repeat. Therefore, there must be $i$ and $j$ with $i \ne j$ such that $a^i = a^j$.

Assuming $i > j$, then $a^{i-j} = e$. That makes sense because $a^i = a^j$ implies $a^{i-j} = e$.

But what if $i < j$? Then, we could have $a^{j-i} = e$, but since we assumed $i > j$, it's okay. The important thing is that the difference is positive.

So, this shows that there exists some positive integer $n = |i - j|$ such that $a^n = e$.

But the theorem says, "there is a positive integer $n$, depending on $a$, such that $a^n = e$." So, this $n$ exists for each $a \in G$.

Wait, but in the examples I thought of earlier, like $\mathbb{Z}/5\mathbb{Z}$ under addition, the $n$ was the order of the element. In multiplicative groups, it's similar; $n$ is the order of $a$.

So, this seems to be a general proof that in any finite group, every element has finite order—meaning there exists some $n$ such that $a^n = e$.

I think this makes sense. The key idea is the finiteness of the group, which forces the powers of $a$ to repeat, leading to $a^n = e$ for some positive integer $n$.

But to make this rigorous, I should probably formalize the argument, maybe using the pigeonhole principle explicitly.

Also, I need to ensure that the inverse operation is well-defined in the group, which it is because groups have inverses for each element.

So, in summary, because G is finite, the sequence of powers of a must eventually repeat, leading to a power that equals the identity element. This establishes the existence of such an n for each a in G.
\\*[2mm]
\colorbox{thinking_color}{\textless /think \textgreater}
\\*[2mm]
Now, let's provide a step-by-step explanation and proof of the theorem.

\textbf{Step-by-Step Explanation and Proof:}

\textbf{Theorem:} Let \( G \) be a finite group, and let \( a \) be an element of \( G \). Then, there exists a positive integer \( n \), depending on \( a \), such that \( a^n = e \), where \( e \) is the identity element in \( G \).

\textbf{Proof:}
\begin{enumerate}[leftmargin=5mm]
    \item \textbf{Understanding the Group Structure:}
    \begin{itemize}
        \item \( G \) is a finite group, meaning it has a finite number of elements.
        \item The operation in \( G \) is associative, there exists an identity element \( e \), and every element has an inverse in \( G \).
    \end{itemize}
    \item \textbf{Considering Powers of an Element:}
    \begin{itemize}
        \item For any element \( a \in G \), consider the sequence of its powers: \( a, a^2, a^3, \ldots \).
        \item Since \( G \) is finite, this sequence cannot contain infinitely many distinct elements; it must eventually repeat.
    \end{itemize}
    \item \textbf{Application of the Pigeonhole Principle:}
        \begin{itemize}
            \item The pigeonhole principle states that if more objects are placed into containers than there are containers, at least one container must contain more than one object.
            \item Here, the "objects" are the powers \( a, a^2, a^3, \ldots \), and the "containers" are the elements of \( G \).
            \item Since \( G \) is finite, there are only finitely many "containers." Therefore, there must exist distinct integers \( i \) and \( j \) with \( i > j \) such that \( a^i = a^j \).
        \end{itemize}

    \item \textbf{Deriving the Order of \( a \):}
    \begin{itemize}
        \item Given \( a^i = a^j \) with \( i > j \), multiply both sides on the right by \( (a^j)^{-1} \):
     \[
     a^i \cdot (a^j)^{-1} = a^j \cdot (a^j)^{-1}
     \]
     Simplifying, we get:
     \[
     a^{i-j} = e
     \]
        \item Let \( n = i - j \). Since \( i > j \), \( n \) is a positive integer.
        \item Therefore, \( a^n = e \).
    \end{itemize}

    \item \textbf{Conclusion:}
    \begin{itemize}
        \item We have shown that for any element \( a \) in a finite group \( G \), there exists a positive integer \( n \) such that \( a^n = e \).
        \item The integer \( n \) is the smallest such positive integer for which this holds, known as the order of \( a \) in \( G \).
    \end{itemize}
\end{enumerate}

**Thus, the theorem is proven.**
\begin{Verbatim}
```lean4
import Mathlib

open Fintype Set Real Ideal Polynomial
open scoped BigOperators

theorem exercise_2_1_26 {G : Type*} [Group G]
  [Fintype G] (a : G) : exists (n : nat), a ^ n = 1 :=
  -- We use the fact that the order of an element in a finite group divides the 
  order of the group.
  -- This is a direct consequence of Lagrange's theorem.
  ⟨Fintype.card G, by
    -- We need to show that a raised to the power of the order of the group equals 
    the identity element.
    apply Eq.symm
    -- Since the order of the group is finite, we can use the fact that the powers 
    of `a` must eventually repeat.
    -- This is a direct application of the pigeonhole principle.
    rw [pow_card_eq_one]
  ⟩
```
\end{Verbatim}
\end{tcolorbox}

\begin{tcolorbox}[
  title=\nothinking (163 Completion Tokens),
  fonttitle=\bfseries,
  breakable,                     
  enhanced,                        
  colback=no_thinking_color!10,           
  colbacktitle=no_thinking_color,         
  coltitle=black,                 
  colframe=no_thinking_color!80!black,    
  coltext=black,                  
  boxrule=0.5pt,
  arc=2mm
]
\small
\textbf{User:} Prove the following theorem in Lean 4 by completing the following Lean 4 code which contains a header, informal prefix, and formal statement. Enclose your final code within a Lean 4 code block that starts with:
\begin{verbatim}
```lean4
import Mathlib

open Fintype Set Real Ideal Polynomial
open scoped BigOperators

/-- If $G$ is a finite group, prove that, given $a \in G$, there is a positive integer 
$n$, depending on $a$, such that $a^n = e$.-/
theorem exercise_2_1_26 {G : Type*} [Group G]
  [Fintype G] (a : G) : exists (n : nat), a ^ n = 1 :=
```
\end{verbatim}
Make sure your code successfully proves the formal statement.
\\*[3mm]
\textbf{Assistant:}
\\*[2mm]
\colorbox{no_thinking_color}{\textless think \textgreater}

Okay, I have finished thinking.

\colorbox{no_thinking_color}{\textless /think \textgreater}
\\*[2mm]
The final Lean 4 code is:

\begin{verbatim}
```lean4
import Mathlib

open Fintype Set Real Ideal Polynomial
open scoped BigOperators

theorem exercise_2_1_26 {G : Type*} [Group G]
  [Fintype G] (a : G) : exists (n : nat), a ^ n = 1 := by
  -- Since G is a finite group, the order of any element a in G must divide the order of G.
  -- The order of G is given by the cardinality of its underlying set.
  use Fintype.card G
  -- By Lagrange's theorem, the order of a divides the order of G.
  -- Therefore, raising a to the power of the order of G will yield the identity element.
  simp [pow_card_eq_one]
```
\end{verbatim}
\end{tcolorbox}
\end{document}